\documentclass[11pt]{article}
\usepackage[final]{acl}

\usepackage{times}
\usepackage{float}
\usepackage{framed}
\usepackage{helvet}
\usepackage{booktabs}
\usepackage{amssymb}
\usepackage{wrapfig}
\usepackage[most]{tcolorbox}
\usepackage{caption}
\usepackage[dvipsnames,svgnames]{xcolor}
\usepackage{amsmath}
\usepackage[T1]{fontenc}
\definecolor{lightgreen}{RGB}{180,220,180}
\definecolor{green}{RGB}{0,150,0}
\definecolor{darkgreen}{RGB}{0,100,0}
\definecolor{shadecolor}{RGB}{230, 240, 255}
\usepackage{latexsym}
\usepackage[utf8]{inputenc}
\usepackage{microtype}
\usepackage{inconsolata}
\usepackage{xcolor}
\usepackage{graphicx}
\usepackage{xspace}
\usepackage{textcomp}

\newcommand{\xlmr}{\textsc{XLM-R}\ensuremath{_{\text{base}}}\xspace}
\newcommand{\xlmrdeen}{\textsc{XLM-R}\textsubscript{de-en}\xspace}
\newcommand{\xlmrenkk}{\textsc{XLM-R}\textsubscript{en-kk}\xspace}
\newcommand{\xlmrjakk}{\textsc{XLM-R}\textsubscript{ja-kk}\xspace}
\newcommand{\xlmrkkko}{\textsc{XLM-R}\textsubscript{kk-ko}\xspace}

\title{Why Better Cross-Lingual Alignment Fails for Better Cross-Lingual Transfer: Case of Encoders}

\author{
Yana Veitsman\textsuperscript{1}\thanks{Work done while at Saarland University. Correspondence to: \texttt{yana.veitsman@uni-goettingen.de}.}
\quad
Yihong Liu\textsuperscript{2,3}
\quad
Hinrich Schütze\textsuperscript{2,3}
\\[0.6em]
\textsuperscript{1}University of Göttingen
\qquad
\textsuperscript{2}LMU Munich\\
\textsuperscript{3}Munich Center for Machine Learning (MCML)
}

\begin{document}
\maketitle
\begin{abstract}
Cross-lingual alignment is often assumed to improve cross-lingual transfer by bringing representations of different languages closer together.
However, improvements in representational alignment do not consistently translate into better downstream performance.
We investigate this disconnect using XLM-R models explicitly aligned across four language pairs with \emph{token-level}, \emph{sentence-level}, and \emph{masked-language-modeling} objectives. 
We evaluate their zero-shot transfer on a token-level task (\emph{part-of-speech tagging}) and a sentence-level task (\emph{sentence classification}), and analyze both representational changes and the gradients induced by the alignment and downstream objectives. 
We find that embedding-based alignment metrics do not reliably indicate whether alignment will improve or degrade downstream performance. 
Moreover, alignment and downstream-task gradients are often nearly \emph{orthogonal}, particularly when the alignment objective and downstream task operate at different representational levels.
These findings suggest that representation alignment alone is insufficient for assessing cross-lingual transfer, and that the compatibility between alignment and downstream objectives should be considered when designing/evaluating alignment methods.
\end{abstract}

\section{Introduction}
Despite their growing multilingual use, large language models (LLMs), including multilingual LLMs, remain heavily English-centric in training data \citep{blasi-etal-2022-systematic, xu2024surveymultilinguallargelanguage, multilingual-survey}. 
Recent multilingual corpora \citep{nguyen-etal-2024-culturax, penedo2024finewebdatasetsdecantingweb, jumelet-etal-2026-babybabellm} and model families \citep{ustun-etal-2024-aya, eurollm, apertus} have expanded linguistic coverage, yet performance remains uneven, particularly for low-resource languages, due to limited training data \citep{veitsman-hartmann-2025-recent}, typological variation \citep{petrov-tokenizers-unfairness}, and capacity constraints such as the \emph{curse of multilinguality} \citep{curse-multilinguality}.

A central strategy for improving performance in low-resource languages is \emph{cross-lingual transfer}, in which knowledge learned from one or more source languages is transferred to a target language under zero-shot or few-shot conditions \citep{nooralahzadeh-etal-2020-zero, asai-etal-2024-buffet}. One widely studied approach to facilitating this transfer is \emph{cross-lingual alignment}, which aims to make representations comparable across languages \citep{wu-dredze-2020-explicit, gaschi-etal-2023-exploring, hammerl-etal-2024-understanding, liu-etal-2025-transliterations}. Initially developed for multilingual encoder models, this principle has motivated a broad range of alignment objectives and training methods \citep{chi-etal-2021-infoxlm, philippy-etal-2023-towards, xhelili-etal-2024-breaking, hammerl-etal-2024-understanding}.

\begin{figure*}[!ht]
    \centering
    \includegraphics[width=0.9\textwidth]{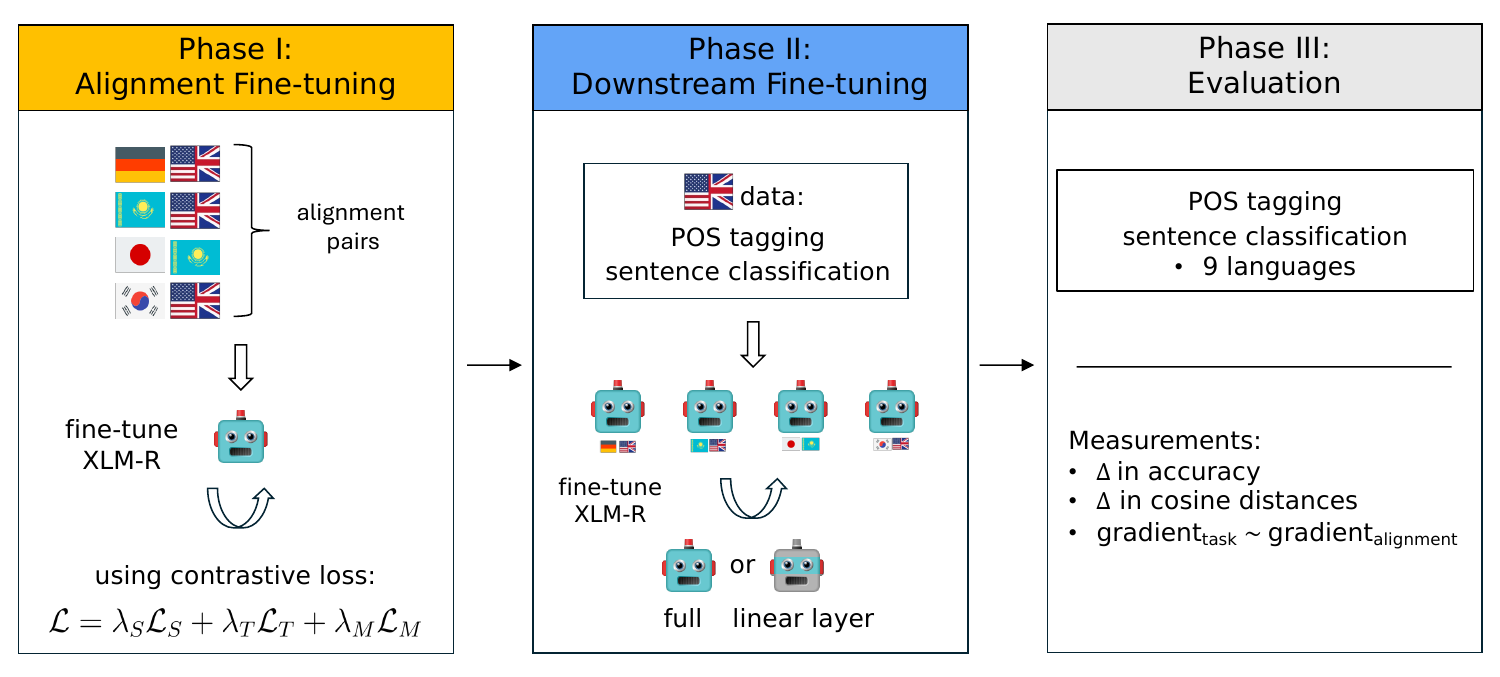}
    \caption{Overview of the experimental setup, including three phases: 1) fine-tuning alignment of \xlmr model on 4 language pairs using a contrastive loss objective; 2) full or linear layer-only fine-tuning downstream task performance of aligned models using English data for \textsc{POS Tagging} and \textsc{Sentence Classification}; 3) evaluation of the models via $\Delta$ in accuracies on the task (zero-shot vs. fine-tuned), $\Delta$ in cosine distances between language embeddings, and alignment and task cosine similarity analysis.}
    \label{fig:graph-abs}
\end{figure*}

Despite its intuitive appeal, the relationship between representational alignment and downstream cross-lingual transfer remains unclear. 
Explicit alignment can substantially improve intrinsic measures such as embedding similarity and cross-lingual retrieval, yet these improvements do not consistently translate into better downstream performance. 
In some cases, alignment even degrades performance on tasks such as part-of-speech tagging and named entity recognition \citep{gaschi-etal-2023-exploring, liu-etal-2024-translico}. 
This raises a fundamental question: \emph{why does better cross-lingual alignment not reliably lead to better cross-lingual transfer?}

\paragraph{Approach and Contributions} We investigate this question from two complementary perspectives. 
First, we examine how explicit alignment changes the multilingual representation space and whether these changes correspond to downstream performance. 
Second, we study the interaction between the optimization objectives  themselves by comparing gradients induced by alignment and downstream-task supervision. 
Intuitively, even if an alignment objective brings representations closer across languages, the resulting parameter updates may not be compatible with those required by a downstream task. 
This distinction allows us to separate \emph{representational proximity} from \emph{optimization compatibility}. For an overview of our approach, refer to Figure \ref{fig:graph-abs}.

Our contributions therefore lie along two axes:
\begin{enumerate}
\item We show that improvements in layer-wise cosine-distance and retrieval-based alignment metrics are not reliable predictors of downstream task performance (Section~\ref{sec:emb-dist}).
\item We find that gradients induced by alignment and task-specific objectives are largely orthogonal, and that their interaction can conflict with the optimization required for improved downstream performance (Section~\ref{sec:gradient-sim}).
\end{enumerate}

Overall, our findings demonstrate that the downstream consequences of alignment are better characterized by optimization compatibility---measured here through gradient similarity---than by representational proximity alone.

\section{Background and Related Work}
We focus on encoder-only Transformer architectures, such as BERT \citep{bert-paper} and XLM-R \citep{curse-multilinguality}. Their cross-lingual generalization is primarily attributed to shared parameters inducing a joint multilingual space \citep{pires-etal-2019-multilingual, dufter-schutze-2020-identifying, KWMR20}. Contextual representations of this space provide a common ground for measuring and defining cross-lingual alignment \citep{hammerl-etal-2024-understanding}.

\paragraph{Cross-Lingual Alignment}
Cross-lingual alignment refers to the extent to which semantically related inputs from different languages occupy similar regions of this shared space. Two notions, commonly referred to as \textit{strong} and \textit{weak} alignment, can be distinguished \citep{roy-etal-2020-lareqa, hammerl-etal-2024-understanding}. Throughout this paper, we use the definition of weak alignment provided in Appendix~\ref{app_a_alignment_def}. 

Alignment can be encouraged at the \textit{token level}, the \textit{sentence level}, or both \citep{hammerl-etal-2024-understanding}. Token-level alignment is particularly relevant to tasks such as part-of-speech tagging and named entity recognition, whereas sentence-level alignment is more directly relevant to sentence classification and cross-lingual retrieval.

\paragraph{Internal Representations}
Languages occupy a shared embedding space, with deeper layers exhibiting stronger cross-lingual alignment \citep{mousi-shared}. Shared tokens may serve as representational anchors \citep{mothello}, and alignment is stronger among related languages, consistent with semantic hubness patterns \citep{wu-semantic-hubness}.

\paragraph{Contrastive Alignment}
We focus on improving explicit representation alignment \citep{wu-dredze-2020-explicit, hu-etal-2021-explicit} via contrastive learning \citep{wei2021on, chi-etal-2021-infoxlm}. Contrastive objectives reduce the distance between representations of semantically equivalent inputs (positive pairs) while increasing their distance from unrelated inputs (negative pairs). Applied at the token and sentence levels, they encourage more consistent cross-lingual representations and a more isotropic embedding space. For example, InfoXLM \citep{chi-etal-2021-infoxlm} maximizes cross-lingual mutual information alongside MLM, while HiCTL \citep{wei2021on} learns representations at the word, sentence, and context levels. In our work, we use a modification of the HiCTL objective.

\paragraph{Alignment and Transfer}
The link between alignment and downstream transfer remains insufficiently explained. Explicit contrastive alignment yields inconsistent gains when controlling for model size and seed variance \citep{wu-dredze-2020-explicit}. Encoder-only models show stronger correlations \citep{gaschi-etal-2023-exploring}, but task-specific fine-tuning can reduce alignment, particularly for token-level tasks, possibly due to catastrophic forgetting. Moreover, stronger alignment does not reliably improve downstream performance and can even degrade it, e.g., with transliteration-based objectives \citep{liu-etal-2024-translico, xhelili-etal-2024-breaking}.

\section{Experimental Setup}
\label{sec:eval}
To study how better cross-lingual alignment interacts with the downstream task performance in multilingual encoders, we develop two orthogonal hypotheses, related to either linguistic typology or to downstream task interaction.

\paragraph{H\textsubscript{I}: Linguistic Relatedness}
Building on typological evidence, we expect that zero-shot transfer depends on the linguistic proximity between (a) alignment language pair, (b) the fine-tuning language, and (c) the target language:
\begin{tcolorbox}[colback=gray!10,colframe=gray!50,title=Hypothesis I]
Zero-shot performance depends on the degree of linguistic relatedness among alignment, fine-tuning, and target languages.
\end{tcolorbox}

\paragraph{H\textsubscript{II}: Competing Objectives}
Previously, \citet{wang-etal-2023-gradsim} reported interference between the gradients of different languages (both typologically similar and distinct) during pre-training stage. We hypothesize that such misalignment can exist between languages, but also between training objectives themselves. Therefore, we formulate the second hypothesis as follows:
\begin{tcolorbox}[colback=gray!10,colframe=gray!50,title=Hypothesis II]
Alignment and downstream-task objectives induce parameter updates in opposing directions, limiting joint optimization.
\end{tcolorbox}

\begin{table*}[!ht]
  \centering
  \small
  \resizebox{\textwidth}{!}{
  \begin{tabular}{lccccccc}
    \toprule
        \multicolumn{8}{c}{\textbf{POS Tagging}} \\
    \midrule
    \multicolumn{8}{c}{Full Model Fine-tuning} \\
    \midrule
    Model & S0T0M1 & S0T1M0 & S0T1M1 & S1T0M0 & S1T0M1 & S1T1M0 & S1T1M1 \\
    \midrule
      \textbf{\xlmr} & $\phantom{-}.72$ & $\phantom{-}.72$ & $\phantom{-}.72$ & $\phantom{-}.72$ & $\phantom{-}.72$ & $\phantom{-}.72$ & $\phantom{-}.72$ \\
    \midrule
    \xlmrdeen & \
    \textcolor{green}{$ \phantom{-}.01 \pm .00 $} & 
    $\phantom{-}.00 \pm .02 $ & 
    $ \phantom{-}.01 \pm .01 $ & 
    $ \phantom{-}.01 \pm .02 $ & 
    \textcolor{green}{$ \phantom{-}.02 \pm .01 $} & 
    $ \phantom{-}.01 \pm .01 $ & 
    \textcolor{green}{$ \phantom{-}.02 \pm .01 $} \\
    \xlmrenkk & \textcolor{red}{$-.02 \pm .01$} & \textcolor{red}{$ -.03 \pm .02 $} & \textcolor{red}{$ -.02 \pm .01 $} & $ \phantom{-}.01 \pm .02 $ & $ \phantom{-}.00 \pm .01 $ & $ \phantom{-}.01 \pm .02 $ & $ \phantom{-}.00 \pm .03 $ \\
    \xlmrjakk & $ \phantom{-}.02 \pm .02 $ & $\phantom{-}.00 \pm .01 $ & \textcolor{green}{$ \phantom{-}.02 \pm .01 $} & $ \phantom{-}.02 \pm .02 $ & \textcolor{green}{$ \phantom{-}.03 \pm .01 $} & $ \phantom{-}.01 \pm .02 $ & \textcolor{green}{$ \phantom{-}.04 \pm .01 $} \\
    \xlmrkkko & $ \phantom{-}.01 \pm .02 $ & $ \phantom{-}.01 \pm .03 $ & $ \phantom{-}.01 \pm .01 $ & $ \phantom{-}.01 \pm .02 $ & $ \phantom{-}.01 \pm .02 $ & $ \phantom{-}.00 \pm .03 $ & $ \phantom{-}.02 \pm .02 $ \\
    \midrule
    \multicolumn{8}{c}{Linear Layer-Only Fine-tuning} \\
    \midrule
    Model & S0T0M1 & S0T1M0 & S0T1M1 & S1T0M0 & S1T0M1 & S1T1M0 & S1T1M1 \\
    \midrule
    \textbf{\xlmr} & \textbf{$\phantom{-}.63$} & \textbf{$\phantom{-}.63$} & \textbf{$\phantom{-}.63$} & \textbf{$\phantom{-}.63$} & \textbf{$\phantom{-}.63$} & \textbf{$\phantom{-}.63$} & \textbf{$\phantom{-}.63$} \\
      \midrule
    \xlmrdeen & \textcolor{green}{$ \phantom{-}.10 \pm .00 $} & $\phantom{-}.01 \pm .01 $ & \textcolor{green}{$ \phantom{-}.08 \pm .00 $} & \textcolor{green}{$ \phantom{-}.02 \pm .00 $} & \textcolor{green}{$ \phantom{-}.04 \pm .00 $} & \textcolor{green}{$ \phantom{-}.02 \pm .01 $} & 
    \textcolor{red}{$ -.03 \pm .00 $} \\
    \xlmrenkk & \textcolor{green}{$ \phantom{-}.08 \pm .00 $} & \textcolor{red}{$ -.09 \pm .00 $} & \textcolor{green}{$ \phantom{-}.06 \pm .00 $} & \textcolor{green}{$ \phantom{-}.01 \pm .00 $} & \textcolor{green}{$ \phantom{-}.07 \pm .00 $} & \textcolor{green}{$ \phantom{-}.03 \pm .00 $} &  \textcolor{red}{$-.01 \pm .00 $} \\
    \xlmrjakk & \textcolor{green}{$ \phantom{-}.10 \pm .00 $} & \textcolor{red}{$ -.02 \pm .00 $} & \textcolor{green}{$ \phantom{-}.08 \pm .00 $} & \textcolor{green}{$ \phantom{-}.02 \pm .00 $} & \textcolor{green}{$ \phantom{-}.03 \pm .00 $} & \textcolor{red}{$ -.02 \pm .00 $} & 
    \textcolor{red}{$ -.04 \pm .00 $} \\
    \xlmrkkko & \textcolor{green}{$ \phantom{-}.06 \pm .00 $} & \textcolor{red}{$ -.02 \pm .00 $} & \textcolor{green}{$ \phantom{-}.06 \pm .00 $} & \textcolor{green}{$ \phantom{-}.04 \pm .00 $} & \textcolor{green}{$ \phantom{-}.01 \pm .00 $} & \textcolor{green}{$ \phantom{-}.02 \pm .00 $} & 
    \textcolor{red}{$ -.02 \pm .00 $} \\
    \midrule \midrule
    
    \multicolumn{8}{c}{\textbf{Sentence Classification}} \\
    \midrule
    \multicolumn{8}{c}{Full Model Fine-tuning} \\
    \midrule
    Model & S0T0M1 & S0T1M0 & S0T1M1 & S1T0M0 & S1T0M1 & S1T1M0 & S1T1M1 \\
    \midrule
    \textbf{\xlmr} & $\phantom{-}.88$ & $\phantom{-}.88$ & $\phantom{-}.88$ & $\phantom{-}.88$ & $\phantom{-}.88$ & $\phantom{-}.88$ & $\phantom{-}.88$ \\ \hline
    \xlmrdeen & $ \phantom{-}.00 \pm .00 $ & \textcolor{red}{$ -.02 \pm .01 $} & $ \phantom{-}.00 \pm .01 $ & $ \phantom{-}.00 \pm .00 $ & \textcolor{green}{$ \phantom{-}.00 \pm .00 $} & \textcolor{red}{$ -.01 \pm .01 $} & \textcolor{green}{$ \phantom{-}.01 \pm .00 $} \\
    \xlmrenkk & $ \phantom{-}.00 \pm .00 $ & \textcolor{red}{$ -.02 \pm .01 $} & $ \phantom{-}.00 \pm .01 $ & $ \phantom{-}.00 \pm .00 $ & \textcolor{green}{$ \phantom{-}.01 \pm .01 $} & \textcolor{red}{$ -.01 \pm .01 $} & $ \phantom{-}.00 \pm .01 $ \\
    \xlmrjakk & $ \phantom{-}.00 \pm .00 $ & \textcolor{red}{$ -.03 \pm .02 $} & $ \phantom{-}.00 \pm .01 $ & \textcolor{red}{$ -.01 \pm .00 $} & $ \phantom{-}.00 \pm .00 $ & $ \phantom{-}.00 \pm .00 $ & $ \phantom{-}.00 \pm .01 $ \\
    \xlmrkkko & $ \phantom{-}.01 \pm .01 $ & \textcolor{red}{$ -.04 \pm .02 $} & $ \phantom{-}.00 \pm .00 $ & $ \phantom{-}.00 \pm .00 $ & $ \phantom{-}.00 \pm .01 $ & \textcolor{red}{$ -.01 \pm .00 $} & \textcolor{red}{$ -.01 \pm .00 $} \\
    \midrule
    \multicolumn{8}{c}{Linear Layer-Only Fine-tuning} \\
    \midrule
    Model & S0T0M1 & S0T1M0 & S0T1M1 & S1T0M0 & S1T0M1 & S1T1M0 & S1T1M1 \\
    \midrule
    \textbf{\xlmr} & $\phantom{-}.72$ & $\phantom{-}.72$ & $\phantom{-}.72$ & $\phantom{-}.72$ & $\phantom{-}.72$ & $\phantom{-}.72$ & $\phantom{-}.72$ \\ \hline
    \xlmrdeen & $ \phantom{-}.01 \pm .01 $ & \textcolor{red}{$ -.09 \pm .00 $} & \textcolor{red}{$ -.05 \pm .01 $} & \textcolor{green}{$ \phantom{-}.15 \pm .00 $} & \textcolor{green}{$ \phantom{-}.15 \pm .00 $} &  \textcolor{green}{$\phantom{-}.14 \pm .00 $} & \textcolor{green}{$ \phantom{-}.13 \pm .00 $} \\
    \xlmrenkk & \textcolor{red}{$ -.04 \pm .00 $} & \textcolor{red}{$ -.20 \pm .01 $} & \textcolor{red}{$ -.04 \pm .01 $} & \textcolor{green}{$ \phantom{-}.10 \pm .00 $} & 
    \textcolor{green}{$ \phantom{-}.11 \pm .00 $} & 
    \textcolor{green}{$ \phantom{-}.10 \pm .00 $} & 
    \textcolor{green}{$ \phantom{-}.10 \pm .00 $} \\
    \xlmrjakk & $ \phantom{-}.02 \pm .01 $ & \textcolor{red}{$ -.06 \pm .01 $} & \textcolor{red}{$ -.06 \pm .00 $} & \textcolor{green}{$ \phantom{-}.11 \pm .00 $} & \textcolor{green}{$ \phantom{-}.12 \pm .00 $} & \textcolor{green}{$ \phantom{-}.10 \pm .00 $} & \textcolor{green}{$ \phantom{-}.12 \pm .00 $} \\
    \xlmrkkko & \textcolor{red}{$ -.02 \pm .01 $} & \textcolor{red}{$ -.18 \pm .00 $} & $ \phantom{-}.01 \pm .01 $ & \textcolor{green}{$ \phantom{-}.13 \pm .00 $} & \textcolor{green}{$ \phantom{-}.11 \pm .00 $} & \textcolor{green}{$ \phantom{-}.11 \pm .00 $} & \textcolor{green}{$ \phantom{-}.11 \pm .00 $} \\
    \bottomrule
  \end{tabular}
  }
  \caption{\textsc{POS Tagging} (\textbf{top)} and \textsc{Sentence Classification} (\textbf{bottom}) results per aligned model and loss configuration. Each block reports $\Delta$ accuracy from \xlmr, averaged over 3 seeds. \textcolor{red}{Red} indicates deterioration, \textcolor{green}{green} indicates improvement, and black indicates inconclusive results.}
  \label{tab:pos_sent_results}
\end{table*}

To evaluate the effects of improved representational alignment, we also test the following propositions suggested by prior work: (i) embedding distances alone are insufficient predictors of task performance; (ii) fine-tuning consistently improves downstream task performance; and (iii) different representation levels capture distinct linguistic properties relevant to different tasks. To examine the above stated hypotheses and propositions systematically, our experimental setup consists of three phases: (i) alignment fine-tuning, (ii) downstream task fine-tuning, and (iii) zero-shot evaluation.

\subsection{Phase I: Cross-lingual Alignment Fine-Tuning}
\paragraph{Model, Data, and Languages}
For all experiments, we use the 270M \xlmr \citep{curse-multilinguality} model. We perform cross-lingual alignment fine-tuning using OPUS \citep{tiedemann-nygaard-2004-opus} or MultiCCAligned \citep{elkishky_ccaligned_2020} datasets, depending on the specific language corpus availability. To allow for better typological diversity, we select German-English, English-Kazakh, Japanese-Kazakh, and Korean-Kazakh language pairs. For brevity, we refer to each checkpoint by the model-language pair abbreviation, for example \xlmrdeen for German-English (and \xlmrenkk, \xlmrjakk, or \xlmrkkko for other pairs respectively). We report relevant data and training parameters in Appendix~\ref{app_b_hyperparams}.

\paragraph{Fine-tuning Objective}
We adopt the Hierarchical Contrastive Learning objective from \citet{wei2021on} with a number of modifications. The overall loss is a sum of scaled losses of two representation levels ($\mathcal{L}_{S}$ and $\mathcal{L}_{T}$) and the regular MLM objective ($\mathcal{L}_{M}$):

\begin{equation}
\mathcal{L}
= \lambda_{S}\mathcal{L}_{S}
+ \lambda_{T}\mathcal{L}_{T}
+ \lambda_{M}\mathcal{L}_{M}
\end{equation}

\paragraph{Sentence-level Loss ($\mathcal{L}_{S}$)} aims to pull sentence of similar semantic content together, while pushing non-similar sentences apart. It is based on the standard InfoNCE loss \cite{oord2019representationlearningcontrastivepredictive}:

\begin{equation}
\mathcal{L_S}=\frac{-1}{2B}\sum_{k=1}^{2B}\log
\frac{\exp(s(z_k,z_{pos(k)}))}
{\sum_{j\neq k}\exp(s(z_k,z_j))},
\end{equation}
where \(B\) is the batch size, \(k\) is an index of the sentence embedding in the parallel dataset, \(z_k\) is the \(k\)-th embedding of the concatenated tensor, \(pos(k)\) is the index of the positive partner of the sentence, and \(s\) is the chosen similarity metric (which is, in our case, cosine similarity).

\paragraph{Token-level ($\mathcal{L}_{T}$) Loss} assumes that tokens often occur in similar contexts. So, it uses a token-level bag-of-words $\mathcal{B}$ from concatenated parallel sentences as a positive set and $\mathcal{S} \subset \mathcal{V} \setminus \mathcal{B}$, where $\mathcal{V}$ is the model vocabulary, as a negative token subset:

{\tiny
\begin{equation}
\mathcal{L}_{T}
= \frac{-1}{|\mathcal{B}|}\!
\sum_{t=1}^{|\mathcal{B}|}
\log\frac{\exp(s(q_i,e(w_t)))}
{\exp(s(q_i,e(w_t)))+\sum_{w_j\in\mathcal{S}}\exp(s(q_i,e(w_j)))},
\end{equation}
}
where \(e\) is the embedding lookup function, \(q\) is the query token embedding, \(t\) is an index in \(\mathcal{B}\), and \(w_t\) is the respective token at that index, and \(s\) is the chosen similarity metric.

\paragraph{Loss Scaling Factor $\lambda_k$.} To evaluate the effect of each loss component independently, we apply a loss scaling factor as defined in \citet{liu-etal-2024-translico}. We enumerate all non-zero combinations of loss coefficients:
\begin{equation}
\lambda_k\in\{0,1\},\quad \sum_{k}\lambda_k>0
\end{equation}
This yields a total of 7 loss configurations $\mathcal{C}$ per alignment pair:.

{\small
\begin{equation}
\mathcal{C} = (\lambda_S, \lambda_T, \lambda_M) \in \{0,1\}^3: (\lambda_S,\lambda_T,\lambda_M)\neq(0,0,0)
\end{equation}
}
For visual clarity, we denote these configurations in the form of \texttt{S1T0M1}, where letters corresponds to loss type followed by the coefficient applied along with that specific loss.

\paragraph{Sentence Retrieval and Embedding Distances}
To verify the effectiveness of the alignment performed in Phase I, we use two widely adopted measures: \textit{sentence retrieval recall} and \textit{cosine-based embedding distances} \cite{hammerl-etal-2024-understanding}. Sentence retrieval is evaluated on the FLORES-200 parallel dataset \citep{flores200} using mean-pooled representations from the 8\textsuperscript{th} layer, while token- and sentence-level embedding distances are measured using MUSE bilingual dictionaries \citep{muse-dictionary} and FLORES-200 \citep{flores200}, respectively. We compare embeddings under three conditions: (i) the base model (\xlmr), (ii) the aligned models (\xlmrdeen, \xlmrenkk, \xlmrkkko, and \xlmrjakk), and (iii) the downstream fine-tuned models. For this analysis, representations are mean-pooled from the 4\textsuperscript{th}, 8\textsuperscript{th}, and 12\textsuperscript{th} layers. Evaluation across these layers is done so the changes within the layers that are supposed to be more or less language-specific can be easily captured.

\begin{figure*}[t]
    \centering
    \includegraphics[width=0.9\textwidth]{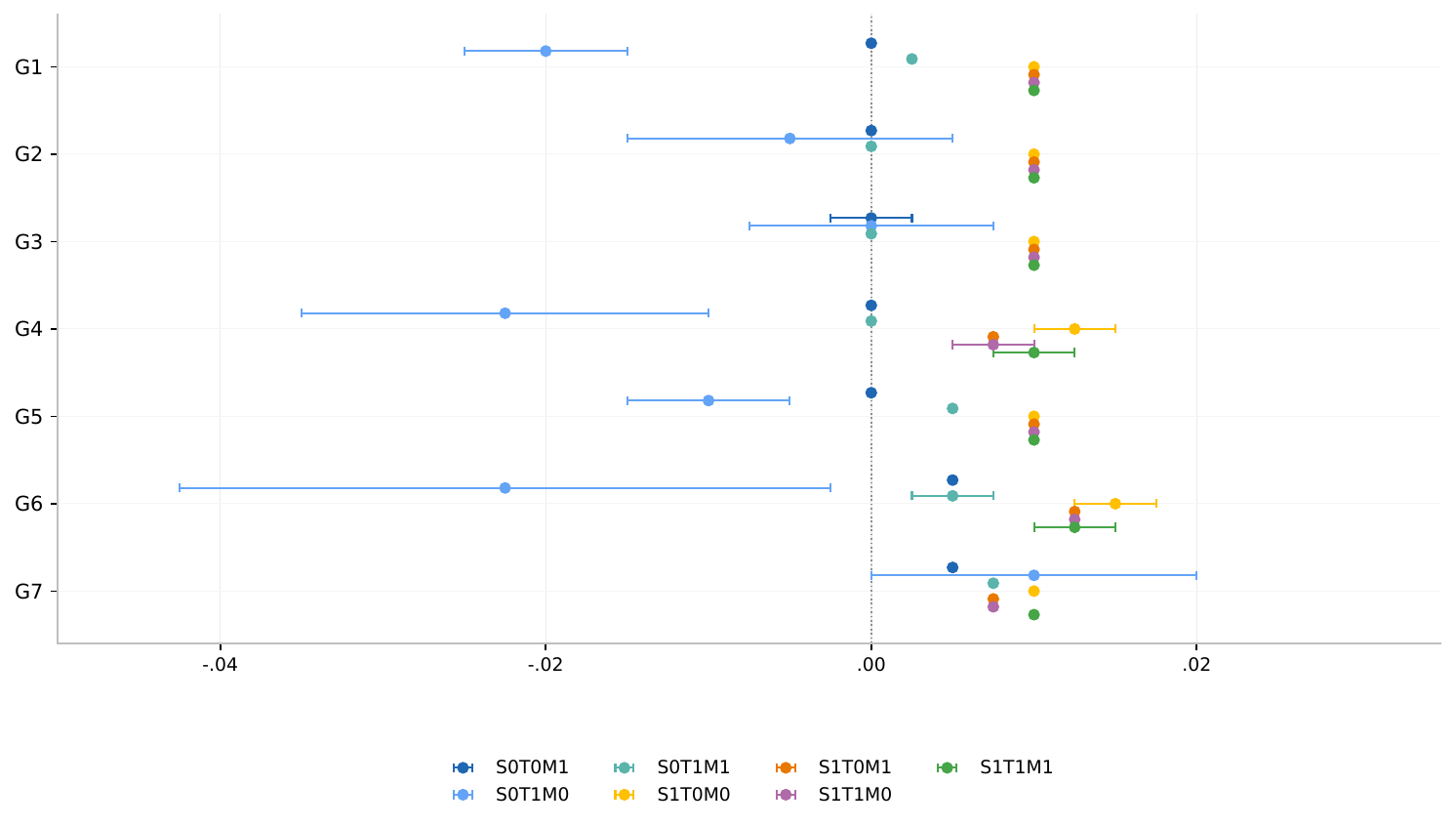}
    \caption{Group-averaged cosine similarity changes across loss configurations. Points show the mean $\Delta$ cosine similarity relative to \xlmr, and whiskers show the averaged standard deviation within each group. Group definitions and detailed results are provided in Appendix, Table \ref{tab:st_similarity_differences}.}
    \label{fig:group_avg}
\end{figure*}

\subsection{Phase II: Downstream Fine-tuning}
After alignment fine-tuning, we separately fine-tune all checkpoints for token-level (\textsc{POS Tagging}) and sentence-level (\textsc{Sentence Classification}) tasks on respective English language data. For \textsc{POS Tagging} task, we use Universal Dependency treebanks \citep{universal-dependencies}.  For \textsc{Sentence Classification}, we employ SIB-200 \citep{sib}, which enhances the FLORES-200 dataset \citep{flores200} with classification labels.

\subsubsection{Fine-tuning Regimen}
Encoder fine-tuning commonly follows one of two strategies: (i) \textit{full fine-tuning}, where the entire model is updated, and (ii) \textit{linear-only fine-tuning}, where only the classification layer is trained while the encoder backbone remains frozen \cite{bert-paper, peters-etal-2019-tune}. This comparison facilitates the investigation of our H\textsubscript{II}. In case the hypothesis is correct, we expect that linear-only fine-tuning will preserve the representations induced by the alignment objective, whereas full fine-tuning would amplify the interference between the objectives.

\paragraph{Gradient Similarity}
Similar to a multi-task learning setup, objectives of alignment and task may produce \textit{conflicting gradients} \citep{conflicting-gradients, wang-etal-2023-gradsim}. Building on this insight, we analyze gradient interactions at both task and language levels. Specifically, we examine (i) gradients between downstream fine-tuning task and alignment objectives, (ii) per-language task gradients after alignment using zero-shot evaluation, and (iii) per-language alignment gradients computed via the MLM objective on FLORES-200 \citep{flores200}.

\subsection{Phase III: Zero-shot Transfer}
After downstream task fine-tuning, we perform zero-shot evaluation of each checkpoint and report the change in classification accuracy for downstream task. The set of evaluation languages for both tasks includes both the languages that were and were not present in the alignment enhancement phase:

{\small
\begin{equation}
L_{\text{POS}} = L_{\text{SENT}}
= \{\text{de},\text{es},\text{fa},\text{fr},\text{hi},\text{ja},\text{kk},\text{ko},\text{ru}\}
\end{equation}
}

\section{Results}
We observe that even though cross-lingual alignment improves after Phase I, this improvement does not directly translate to better performance, depending on the fine-tuning regimen and loss configuration. We attribute this to the gradient conflict between the alignment and the downstream task training objectives.

\subsection{Embedding Distances}
\label{sec:emb-dist}

\begin{figure}[!t]
    \centering
    \includegraphics[width=0.8\columnwidth]{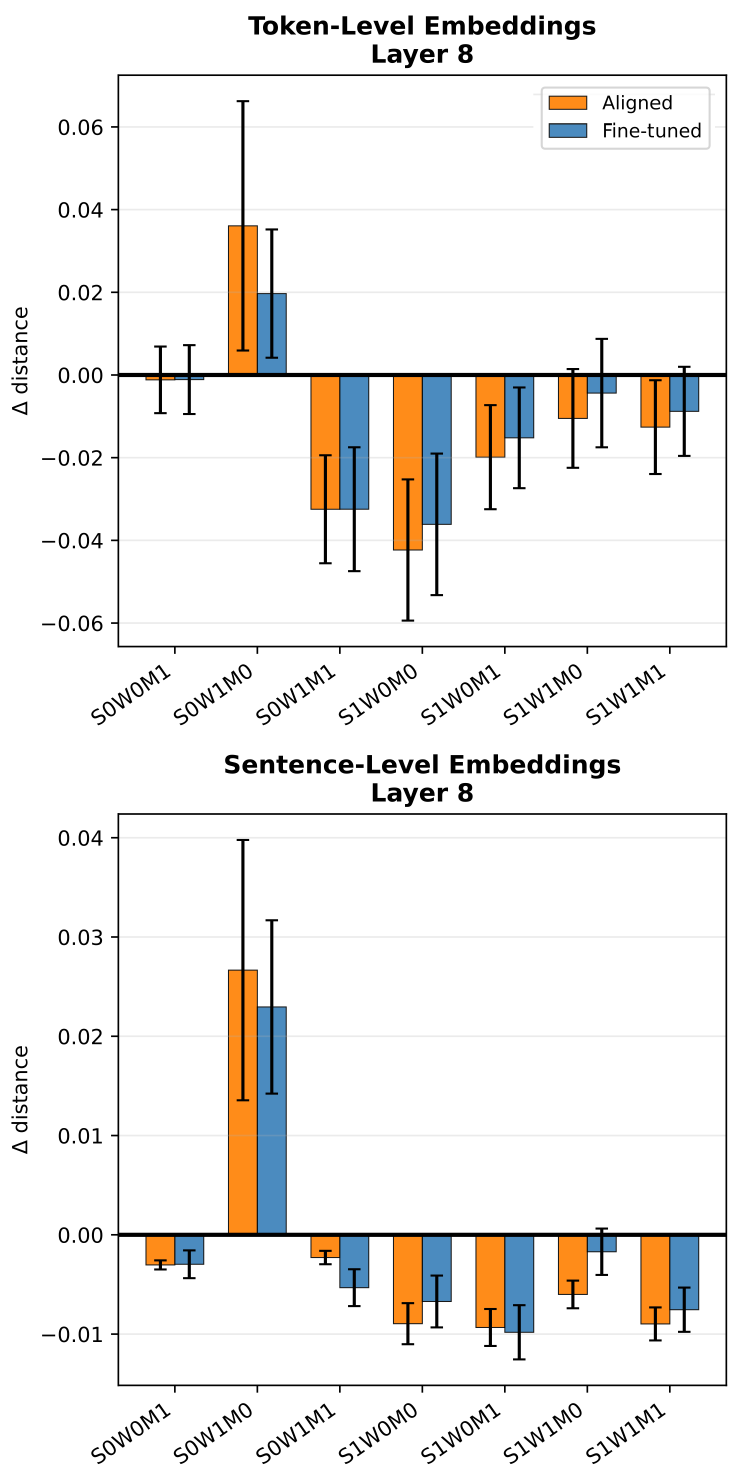}
    \caption{$\Delta$ in cosine distance from \xlmr for embeddings
    averaged across models and languages and grouped by loss configuration. Full figure, including layers 4 and 12, is available in the Figure \ref{fig:cos-dists} in Appendix.}
    \label{fig:dist_embs}
\end{figure}

After Phase I, we evaluate changes in token- and sentence-level embedding distances and sentence-level cosine similarity and retrieval. Despite substantial cross-lingual variance, consistent layer-, model-, and objective-specific patterns emerge. 

\paragraph{Distances Across Layers} Layer-wise results ( Figure~\ref{fig:cos-dists} in Appendix) support prior distinctions between relatively language-agnostic and more language- or task-specific layers \cite{pires-etal-2019-multilingual, tanti-etal-2021-language, choenni-shutova-2022-investigating}. Distances generally decrease in the 8\textsuperscript{th} layer for both aligned and downstream fine-tuned models, except under the $S0T1M0$ configuration (Figure~\ref{fig:dist_embs}). Changes in lower and final layers, which are more heterogeneous, likely reflect language- and task-specific adaptation.

\paragraph{Cosine Similarities} Sentence-level similarity gains are generally modest, partly because some baseline pairs are already highly similar (Figure \ref{fig:group_avg}, details in Appendix, Table \ref{tab:st_similarity_differences}). For example, Spanish--English reaches approximately $.98$. Most improvements remain below $.02$, with the smallest gains observed for non-aligned pairs involving English and different scripts (Group~3: hin--eng, zho--eng, spa--eng, and rus--eng). This pattern suggests that alignment fine-tuning may partially shift these languages away from the English-centered representational space. Sentence retrieval results exhibit the same tendency (see Tables \ref{tab:delta-r5} and \ref{tab:delta-r10} in Appendix).

\subsection{Downstream Task Performance}
We find that downstream task performance is rather driven by the specific loss configuration and fine-tuning regimen, than a particular language alignment pair (Table~\ref{tab:pos_sent_results}).

\paragraph{Loss Configuration} Some loss combinations generally hurt the performance across tasks, languages, and fine-tuning modes. For example, in the case of the $S0T1M0$ configuration, performance degrades in both fine-tuning modes and tasks. On the other hand, configurations that include a combination of sentence-level loss with either or both token-level and MLM objectives, tend to improve the downstream accuracy more often. Moreover, inclusion of sentence-level loss also positively affects \textsc{POS Tagging}, specifically in the linear layer fine-tuning case.

\paragraph{Fine-tuning Regimen} Choice of the fine-tuning regimen is also impactful: full model fine-tuning tends to affect task solving abilities negatively, while the linear layer-only fine-tuning shows more robust improvements.  For example, \xlmrdeen experiences up to $15\%$ accuracy gain on \textsc{Sentence Classification} in the case of $S1T0M0$ and $S1T0M1$ configurations. Consistent improvements are also seen in \textsc{POS Tagging}, \xlmrenkk becomes up to $7\%$ more accurate.

Thus, although the similarity metrics showcase improved alignment, the impact on downstream performance is more fine-grained, particularly so more within the applied loss and training mode, than language pairs.

\subsection{Gradient Similarity}
\label{sec:gradient-sim}
\begin{figure*}
    \centering
    \includegraphics[width=0.9\textwidth]{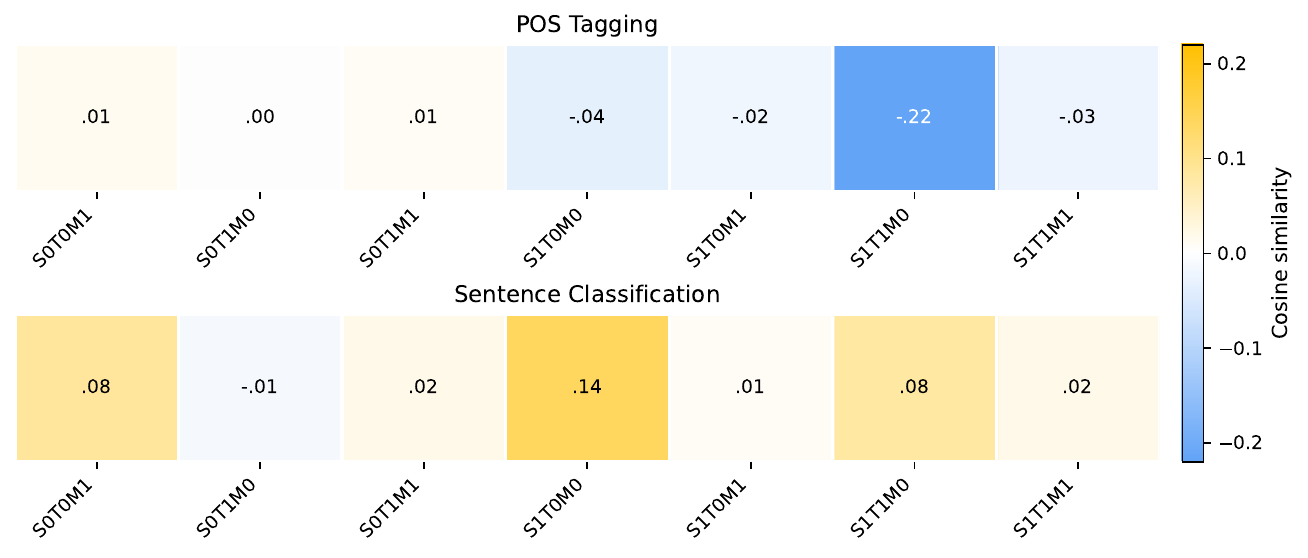}
    \caption{Cosine similarity between gradients induced by the alignment objective and those induced by each downstream task: \textsc{POS Tagging} (\textbf{top}) and \textsc{Sentence Classification} (\textbf{bottom}), averaged over four alignment pairs and grouped by loss configuration. Detailed results per language pair with corresponding standard deviations are available in Appendix, Table~\ref{tab:align_task_grad}.}
    \label{fig:grad-sim-heatmap}
\end{figure*}

To test H\textsubscript{II}, we calculate the cosine similarity between task and alignment objectives. Here we observe a clear separation: gradients for \textsc{Sentence Classification} are generally more positively aligned with the alignment objective, whereas \textsc{POS Tagging} gradients frequently approach zero or become negative, indicating orthogonality or conflict (Figure~\ref{fig:grad-sim-heatmap}). These results suggest fundamental differences in how alignment objectives interact with token- vs. sentence-level supervision.

Language pair and loss configuration refine this picture. Alignment pairs that exclude English show higher gradient similarity and thus reduced task conflict. On the other hand, English-involving alignments exhibit stronger task separation, with consistently positive cosine values for \textsc{Sentence Classification} and negative ones for \textsc{POS Tagging}. Adding a sentence-level alignment loss increases similarity for sentence-level tasks as expected, whereas the MLM-only setup shows moderate synergy with \textsc{Sentence Classification} but remains weakly aligned with token-level objectives. Overall, the findings confirm that alignment and downstream tasks differ substantially in optimization behavior, depending on representation level and alignment configuration.

\section{Discussion}
We confirm that improving cross-lingual representational alignment is not, by itself, sufficient to improve zero-shot downstream performance. Instead, the effect of alignment fine-tuning depends on which representation level is optimized, how compatible the alignment and downstream objectives are, and whether subsequent fine-tuning preserves or modifies the aligned encoder. 

Alignment fine-tuning consistently changes the multilingual representation space, particularly in the 8\textsuperscript{th} layer, where token- and sentence-level distances generally decrease. Sentence-level cosine similarities and retrieval performance also improve for many language pairs. Nevertheless, these representational improvements do not consistently translate into higher downstream accuracy. Some configurations produce clearer cross-lingual similarity while having little effect on task performance, and the token-only \texttt{S0T1M0} objective frequently deteriorates performance despite explicitly optimizing token-level alignment.

\begin{tcolorbox}[colback=gray!10,colframe=gray!50,title=Takeaway I]
Embedding similarity and sentence-retrieval metrics are insufficient predictors of downstream task performance.
\end{tcolorbox}

Gradient analysis provides a more direct explanation for the discrepancy between representational alignment and downstream performance. Gradients from \textsc{Sentence Classification} are more positively aligned with the preceding alignment objective, whereas gradients from \textsc{POS Tagging} are frequently close to orthogonal or negatively aligned. Thus, the two downstream tasks interact differently with the same aligned representation space: sentence-level supervision can often reuse directions encouraged during alignment fine-tuning, while token-level supervision more frequently requires updates that compete with them. This conclusion is supported by the comparison of downstream accuracy results in the fully and linear layer-only tuned models, where tuning across layers seems to amplify the interference.

\begin{tcolorbox}[colback=gray!10,colframe=gray!50,title=Takeaway II]
Compatibility between alignment and downstream objectives depends on the task and representation level.
\end{tcolorbox}

More broadly, the results emphasize that the usefulness of alignment depends not only on how close representations become, but also on which information is preserved and whether the downstream objective can exploit them without undoing the alignment itself.

\section{Conclusion}
We show that improved cross-lingual representational alignment does not necessarily translate into better downstream transfer: embedding similarity can indicate stronger alignment even when downstream performance remains unchanged or deteriorates. 
Our gradient analysis provides a complementary explanation for this disconnect, showing that alignment and downstream objectives are often weakly aligned or conflicting, particularly when they operate at different representational levels. 
We further find that linear-only fine-tuning more consistently preserves the benefits of alignment than full-model fine-tuning. 
Overall, our results suggest that representational alignment should be treated as a diagnostic model property rather than a task-independent proxy for transfer, and highlight the importance of considering the compatibility between alignment and downstream objectives when designing cross-lingual alignment methods.
\section{Limitations}
\paragraph{Model Architecture} Our analysis is constrained by its reliance on a single Transformer-based encoder model, which limits the generality of the findings. Different encoder and decoder-only architectures exhibit varying multilingual capabilities, and extending the study would help test whether the observed dynamics hold more broadly. 
\paragraph{Pre-training and Tokenization Disparities} We do not examine the effect of pre-training language distributions, data quality, or volume on the effectiveness of the cross-lingual alignment. Additionally, we do not account for potential tokenization disparities, which might further drive linguistic inequality.
\paragraph{Training Objective Limitations} We test only one specific contrastive learning objective. Other design choices, for example, replacing the token-level alignment with a carefully designed word-level alignment using bilingual dictionaries, may be of interest for future work. 

\section{Acknowledgements}
YV thanks Vera Demberg, Mark Rofin, and Oleh Bakumenko for useful discussions and feedback on the draft. YL and HS were supported by DFG (grant SCHU 2246/14-1).

\newpage

\bibliography{custom}

@inproceedings{gaschi-etal-2023-exploring,
    title = "Exploring the Relationship between Alignment and Cross-lingual Transfer in Multilingual Transformers",
    author = "Gaschi, Felix  and
      Cerda, Patricio  and
      Rastin, Parisa  and
      Toussaint, Yannick",
    editor = "Rogers, Anna  and
      Boyd-Graber, Jordan  and
      Okazaki, Naoaki",
    booktitle = "Findings of the Association for Computational Linguistics: ACL 2023",
    month = jul,
    year = "2023",
    address = "Toronto, Canada",
    publisher = "Association for Computational Linguistics",
    url = "https://aclanthology.org/2023.findings-acl.189/",
    doi = "10.18653/v1/2023.findings-acl.189",
    pages = "3020--3042"
}

@inproceedings{jumelet-etal-2026-babybabellm,
  title = {{{BabyBabelLM}}: {{A}} Multilingual Benchmark of Developmentally Plausible Training Data},
  shorttitle = {{{BabyBabelLM}}},
  booktitle = {Proceedings of the 19th {{Conference}} of the {{European Chapter}} of the {{Association}} for {{Computational Linguistics}} ({{Volume}} 1: {{Long Papers}})},
  author = {Jumelet, Jaap and Fourtassi, Abdellah and Haga, Akari and Bunzeck, Bastian and Shandilya, Bhargav and {Galvan-Sosa}, Diana and Haznitrama, Faiz Ghifari and Padovani, Francesca and Meyer, Francois and Hu, Hai and Etxaniz, Julen and Prevot, Laurent and He, Linyang and Grandury, Mar{\'i}a and Marcheva, Mila and Foroutan, Negar and Theodoropoulos, Nikitas and Sadeghi, Pouya and Song, Siyuan and Salhan, Suchir and Zhou, Susana and Paniv, Yurii and Zhang, Ziyin and Bisazza, Arianna and Warstadt, Alex and Choshen, Leshem},
  year = 2026,
  pages = {3297--3329},
  publisher = {Association for Computational Linguistics},
  address = {Rabat, Morocco},
  doi = {10.18653/v1/2026.eacl-long.152},
  url = {https://aclanthology.org/2026.eacl-long.152},
  urldate = {2026-06-22},
  langid = {english}
}

@article{eurollm,
title = {EuroLLM: Multilingual Language Models for Europe},
journal = {Procedia Computer Science},
volume = {255},
pages = {53-62},
year = {2025},
note = {Proceedings of the Second EuroHPC user day},
issn = {1877-0509},
doi = {https://doi.org/10.1016/j.procs.2025.02.260},
url = {https://www.sciencedirect.com/science/article/pii/S1877050925006210},
author = {Pedro Henrique Martins and Patrick Fernandes and João Alves and Nuno M. Guerreiro and Ricardo Rei and Duarte M. Alves and José Pombal and Amin Farajian and Manuel Faysse and Mateusz Klimaszewski and Pierre Colombo and Barry Haddow and José G.C. {de Souza} and Alexandra Birch and André F.T. Martins}
}

@inproceedings{hammerl-etal-2024-understanding,
    title = "Understanding Cross-Lingual {A}lignment{---}{A} Survey",
    author = {H{\"a}mmerl, Katharina  and
      Libovick{\'y}, Jind{\v{r}}ich  and
      Fraser, Alexander},
    editor = "Ku, Lun-Wei  and
      Martins, Andre  and
      Srikumar, Vivek",
    booktitle = "Findings of the Association for Computational Linguistics: ACL 2024",
    month = aug,
    year = "2024",
    address = "Bangkok, Thailand",
    publisher = "Association for Computational Linguistics",
    url = "https://aclanthology.org/2024.findings-acl.649/",
    doi = "10.18653/v1/2024.findings-acl.649",
    pages = "10922--10943"
}

@inproceedings{philippy-etal-2023-towards,
    title = "Towards a Common Understanding of Contributing Factors for Cross-Lingual Transfer in Multilingual Language Models: A Review",
    author = "Philippy, Fred  and
      Guo, Siwen  and
      Haddadan, Shohreh",
    editor = "Rogers, Anna  and
      Boyd-Graber, Jordan  and
      Okazaki, Naoaki",
    booktitle = "Proceedings of the 61st Annual Meeting of the Association for Computational Linguistics (Volume 1: Long Papers)",
    month = jul,
    year = "2023",
    address = "Toronto, Canada",
    publisher = "Association for Computational Linguistics",
    url = "https://aclanthology.org/2023.acl-long.323/",
    doi = "10.18653/v1/2023.acl-long.323",
    pages = "5877--5891"
}

@inproceedings{KWMR20,
    author = {Karthikeyan K and Zihan Wang and Stephen Mayhew and Dan Roth},
    title = {{Cross-Lingual Ability of Multilingual BERT: An Empirical Study}},
    booktitle = {Proc. of the International Conference on Learning Representations},
    year = {2020},
    url = "https://cogcomp.seas.upenn.edu/papers/KWMR20.pdf",
    funding = {LORELEI, CwC},
}

@article{xu2024surveymultilinguallargelanguage,
  title={A survey on multilingual large language models: corpora, alignment, and bias},
  author={Yuemei Xu and Ling Hu and Jiayi Zhao and Zihan Qiu and Kexin XU and Yuqi Ye and Hanwen Gu},
  journal={Frontiers of Computer Science},
  volume={19},
  pages={1911362},
  year={2025},
  publisher={Springer},
  doi={10.1007/s11704-024-40579-4},
  url={https://doi.org/10.1007/s11704-024-40579-4}
}

@inproceedings{veitsman-hartmann-2025-recent,
    title = "Recent Advancements and Challenges of {T}urkic {C}entral {A}sian Language Processing",
    author = "Veitsman, Yana  and
      Hartmann, Mareike",
    editor = "Hettiarachchi, Hansi  and
      Ranasinghe, Tharindu  and
      Rayson, Paul  and
      Mitkov, Ruslan  and
      Gaber, Mohamed  and
      Premasiri, Damith  and
      Tan, Fiona Anting  and
      Uyangodage, Lasitha",
    booktitle = "Proceedings of the First Workshop on Language Models for Low-Resource Languages",
    month = jan,
    year = "2025",
    address = "Abu Dhabi, United Arab Emirates",
    publisher = "Association for Computational Linguistics",
    url = "https://aclanthology.org/2025.loreslm-1.25/",
    pages = "309--324"
}

@inproceedings{wu-dredze-2020-explicit,
    title = "Do Explicit Alignments Robustly Improve Multilingual Encoders?",
    author = "Wu, Shijie  and
      Dredze, Mark",
    editor = "Webber, Bonnie  and
      Cohn, Trevor  and
      He, Yulan  and
      Liu, Yang",
    booktitle = "Proceedings of the 2020 Conference on Empirical Methods in Natural Language Processing (EMNLP)",
    month = nov,
    year = "2020",
    address = "Online",
    publisher = "Association for Computational Linguistics",
    url = "https://aclanthology.org/2020.emnlp-main.362/",
    doi = "10.18653/v1/2020.emnlp-main.362",
    pages = "4471--4482"
}

@inproceedings{conflicting-gradients,
author = {Sener, Ozan and Koltun, Vladlen},
title = {Multi-task learning as multi-objective optimization},
year = {2018},
publisher = {Curran Associates Inc.},
address = {Red Hook, NY, USA},
booktitle = {Proceedings of the 32nd International Conference on Neural Information Processing Systems},
pages = {525–536},
numpages = {12},
location = {Montr\'{e}al, Canada}
}

@inproceedings{universal-dependencies,
    title = "{U}niversal {D}ependencies v2: An Evergrowing Multilingual Treebank Collection",
    author = "Nivre, Joakim  and
      de Marneffe, Marie-Catherine  and
      Ginter, Filip  and
      Haji{\v{c}}, Jan  and
      Manning, Christopher D.  and
      Pyysalo, Sampo  and
      Schuster, Sebastian  and
      Tyers, Francis  and
      Zeman, Daniel",
    editor = "Calzolari, Nicoletta  and
      B{\'e}chet, Fr{\'e}d{\'e}ric  and
      Blache, Philippe  and
      Choukri, Khalid  and
      Cieri, Christopher  and
      Declerck, Thierry  and
      Goggi, Sara  and
      Isahara, Hitoshi  and
      Maegaard, Bente  and
      Mariani, Joseph  and
      Mazo, H{\'e}l{\`e}ne  and
      Moreno, Asuncion  and
      Odijk, Jan  and
      Piperidis, Stelios",
    booktitle = "Proceedings of the Twelfth Language Resources and Evaluation Conference",
    month = may,
    year = "2020",
    address = "Marseille, France",
    publisher = "European Language Resources Association",
    url = "https://aclanthology.org/2020.lrec-1.497/",
    pages = "4034--4043",
    language = "eng",
    ISBN = "979-10-95546-34-4"
}

@inproceedings{tanti-etal-2021-language,
    title = "On the Language-specificity of Multilingual {BERT} and the Impact of Fine-tuning",
    author = "Tanti, Marc  and
      van der Plas, Lonneke  and
      Borg, Claudia  and
      Gatt, Albert",
    editor = "Bastings, Jasmijn  and
      Belinkov, Yonatan  and
      Dupoux, Emmanuel  and
      Giulianelli, Mario  and
      Hupkes, Dieuwke  and
      Pinter, Yuval  and
      Sajjad, Hassan",
    booktitle = "Proceedings of the Fourth BlackboxNLP Workshop on Analyzing and Interpreting Neural Networks for NLP",
    month = nov,
    year = "2021",
    address = "Punta Cana, Dominican Republic",
    publisher = "Association for Computational Linguistics",
    url = "https://aclanthology.org/2021.blackboxnlp-1.15/",
    doi = "10.18653/v1/2021.blackboxnlp-1.15",
    pages = "214--227"
}

@inproceedings{roy-etal-2020-lareqa,
    title = "{LAR}e{QA}: Language-Agnostic Answer Retrieval from a Multilingual Pool",
    author = "Roy, Uma  and
      Constant, Noah  and
      Al-Rfou, Rami  and
      Barua, Aditya  and
      Phillips, Aaron  and
      Yang, Yinfei",
    editor = "Webber, Bonnie  and
      Cohn, Trevor  and
      He, Yulan  and
      Liu, Yang",
    booktitle = "Proceedings of the 2020 Conference on Empirical Methods in Natural Language Processing (EMNLP)",
    month = nov,
    year = "2020",
    address = "Online",
    publisher = "Association for Computational Linguistics",
    url = "https://aclanthology.org/2020.emnlp-main.477/",
    doi = "10.18653/v1/2020.emnlp-main.477",
    pages = "5919--5930"
}

@inproceedings{hu-etal-2021-explicit,
    title = "Explicit Alignment Objectives for Multilingual Bidirectional Encoders",
    author = "Hu, Junjie  and
      Johnson, Melvin  and
      Firat, Orhan  and
      Siddhant, Aditya  and
      Neubig, Graham",
    editor = "Toutanova, Kristina  and
      Rumshisky, Anna  and
      Zettlemoyer, Luke  and
      Hakkani-Tur, Dilek  and
      Beltagy, Iz  and
      Bethard, Steven  and
      Cotterell, Ryan  and
      Chakraborty, Tanmoy  and
      Zhou, Yichao",
    booktitle = "Proceedings of the 2021 Conference of the North American Chapter of the Association for Computational Linguistics: Human Language Technologies",
    month = jun,
    year = "2021",
    address = "Online",
    publisher = "Association for Computational Linguistics",
    url = "https://aclanthology.org/2021.naacl-main.284/",
    doi = "10.18653/v1/2021.naacl-main.284",
    pages = "3633--3643"
}

@inproceedings{nooralahzadeh-etal-2020-zero,
    title = "Zero-Shot Cross-Lingual Transfer with Meta Learning",
    author = "Nooralahzadeh, Farhad  and
      Bekoulis, Giannis  and
      Bjerva, Johannes  and
      Augenstein, Isabelle",
    editor = "Webber, Bonnie  and
      Cohn, Trevor  and
      He, Yulan  and
      Liu, Yang",
    booktitle = "Proceedings of the 2020 Conference on Empirical Methods in Natural Language Processing (EMNLP)",
    month = nov,
    year = "2020",
    address = "Online",
    publisher = "Association for Computational Linguistics",
    url = "https://aclanthology.org/2020.emnlp-main.368/",
    doi = "10.18653/v1/2020.emnlp-main.368",
    pages = "4547--4562"
}

@inproceedings{asai-etal-2024-buffet,
    title = "{BUFFET}: Benchmarking Large Language Models for Few-shot Cross-lingual Transfer",
    author = "Asai, Akari  and
      Kudugunta, Sneha  and
      Yu, Xinyan  and
      Blevins, Terra  and
      Gonen, Hila  and
      Reid, Machel  and
      Tsvetkov, Yulia  and
      Ruder, Sebastian  and
      Hajishirzi, Hannaneh",
    editor = "Duh, Kevin  and
      Gomez, Helena  and
      Bethard, Steven",
    booktitle = "Proceedings of the 2024 Conference of the North American Chapter of the Association for Computational Linguistics: Human Language Technologies (Volume 1: Long Papers)",
    month = jun,
    year = "2024",
    address = "Mexico City, Mexico",
    publisher = "Association for Computational Linguistics",
    url = "https://aclanthology.org/2024.naacl-long.100/",
    doi = "10.18653/v1/2024.naacl-long.100",
    pages = "1771--1800"
}

@inproceedings{xhelili-etal-2024-breaking,
    title = "Breaking the Script Barrier in Multilingual Pre-Trained Language Models with Transliteration-Based Post-Training Alignment",
    author = "Xhelili, Orgest  and
      Liu, Yihong  and
      Schuetze, Hinrich",
    editor = "Al-Onaizan, Yaser  and
      Bansal, Mohit  and
      Chen, Yun-Nung",
    booktitle = "Findings of the Association for Computational Linguistics: EMNLP 2024",
    month = nov,
    year = "2024",
    address = "Miami, Florida, USA",
    publisher = "Association for Computational Linguistics",
    url = "https://aclanthology.org/2024.findings-emnlp.659/",
    doi = "10.18653/v1/2024.findings-emnlp.659",
    pages = "11283--11296"
}

@article{flores200,
  author    = {NLLB Team and Marta R. Costa-jussà and James Cross and Onur Çelebi and Maha Elbayadand Kenneth Heafield and Kevin Heffernan and Elahe Kalbassi and  Janice Lam and Daniel Licht and Jean Maillard and Anna Sun and Skyler Wang and Guillaume Wenzek and Al Youngblood and Bapi Akula and Loic Barrault and Gabriel Mejia Gonzalez and Prangthip Hansanti and John Hoffman and Semarley Jarrett and Kaushik Ram Sadagopan and Dirk Rowe and Shannon Spruit and Chau Tran and Pierre Andrews and Necip Fazil Ayan and Shruti Bhosale and Sergey Edunov and Angela Fan and Cynthia Gao and Vedanuj Goswami and Francisco Guzmán and Philipp Koehn and Alexandre Mourachko and Christophe Ropers and Safiyyah Saleem and Holger Schwenk and Jeff Wang},
  title     = {No Language Left Behind: Scaling Human-Centered Machine Translation},
  year      = {2022}
}

@misc{oord2019representationlearningcontrastivepredictive,
      title={Representation Learning with Contrastive Predictive Coding}, 
      author={Aaron van den Oord and Yazhe Li and Oriol Vinyals},
      year={2019},
      eprint={1807.03748},
      archivePrefix={arXiv},
      primaryClass={cs.LG},
      url={https://arxiv.org/abs/1807.03748}, 
}

@inproceedings{elkishky_ccaligned_2020,
 author = {El-Kishky, Ahmed and Chaudhary, Vishrav and Guzm{\'a}n, Francisco and Koehn, Philipp},
 booktitle = {Proceedings of the 2020 Conference on Empirical Methods in Natural Language Processing (EMNLP 2020)},
 month = {November},
 title = {{CCAligned}: A Massive Collection of Cross-lingual Web-Document Pairs},
 year = {2020},
 address = "Online",
 publisher = "Association for Computational Linguistics",
 url = "https://www.aclweb.org/anthology/2020.emnlp-main.480",
 doi = "10.18653/v1/2020.emnlp-main.480",
 pages = "5960--5969"
}

@inproceedings{tiedemann-nygaard-2004-opus,
    title = "The {OPUS} Corpus - Parallel and Free: \url{http://logos.uio.no/opus}",
    author = {Tiedemann, J{\"o}rg  and
      Nygaard, Lars},
    editor = "Lino, Maria Teresa  and
      Xavier, Maria Francisca  and
      Ferreira, F{\'a}tima  and
      Costa, Rute  and
      Silva, Raquel",
    booktitle = "Proceedings of the Fourth International Conference on Language Resources and Evaluation ({LREC}{'}04)",
    month = may,
    year = "2004",
    address = "Lisbon, Portugal",
    publisher = "European Language Resources Association (ELRA)",
    url = "https://aclanthology.org/L04-1174/"
}

@article{multilingual-survey,
title = {A survey of multilingual large language models},
journal = {Patterns},
volume = {6},
number = {1},
pages = {101118},
year = {2025},
issn = {2666-3899},
doi = {https://doi.org/10.1016/j.patter.2024.101118},
url = {https://www.sciencedirect.com/science/article/pii/S2666389924002903},
author = {Libo Qin and Qiguang Chen and Yuhang Zhou and Zhi Chen and Yinghui Li and Lizi Liao and Min Li and Wanxiang Che and Philip S. Yu}
}

@inproceedings{liu-etal-2024-translico,
    title = "{T}ransli{C}o: A Contrastive Learning Framework to Address the Script Barrier in Multilingual Pretrained Language Models",
    author = "Liu, Yihong  and
      Ma, Chunlan  and
      Ye, Haotian  and
      Schuetze, Hinrich",
    editor = "Ku, Lun-Wei  and
      Martins, Andre  and
      Srikumar, Vivek",
    booktitle = "Proceedings of the 62nd Annual Meeting of the Association for Computational Linguistics (Volume 1: Long Papers)",
    month = aug,
    year = "2024",
    address = "Bangkok, Thailand",
    publisher = "Association for Computational Linguistics",
    url = "https://aclanthology.org/2024.acl-long.136/",
    doi = "10.18653/v1/2024.acl-long.136",
    pages = "2476--2499"
}

@inproceedings{blasi-etal-2022-systematic,
    title = "Systematic Inequalities in Language Technology Performance across the World`s Languages",
    author = "Blasi, Damian  and
      Anastasopoulos, Antonios  and
      Neubig, Graham",
    editor = "Muresan, Smaranda  and
      Nakov, Preslav  and
      Villavicencio, Aline",
    booktitle = "Proceedings of the 60th Annual Meeting of the Association for Computational Linguistics (Volume 1: Long Papers)",
    month = may,
    year = "2022",
    address = "Dublin, Ireland",
    publisher = "Association for Computational Linguistics",
    url = "https://aclanthology.org/2022.acl-long.376/",
    doi = "10.18653/v1/2022.acl-long.376",
    pages = "5486--5505"
}

@inproceedings{chi-etal-2021-infoxlm,
    title = "{I}nfo{XLM}: An Information-Theoretic Framework for Cross-Lingual Language Model Pre-Training",
    author = "Chi, Zewen  and
      Dong, Li  and
      Wei, Furu  and
      Yang, Nan  and
      Singhal, Saksham  and
      Wang, Wenhui  and
      Song, Xia  and
      Mao, Xian-Ling  and
      Huang, Heyan  and
      Zhou, Ming",
    editor = "Toutanova, Kristina  and
      Rumshisky, Anna  and
      Zettlemoyer, Luke  and
      Hakkani-Tur, Dilek  and
      Beltagy, Iz  and
      Bethard, Steven  and
      Cotterell, Ryan  and
      Chakraborty, Tanmoy  and
      Zhou, Yichao",
    booktitle = "Proceedings of the 2021 Conference of the North American Chapter of the Association for Computational Linguistics: Human Language Technologies",
    month = jun,
    year = "2021",
    address = "Online",
    publisher = "Association for Computational Linguistics",
    url = "https://aclanthology.org/2021.naacl-main.280/",
    doi = "10.18653/v1/2021.naacl-main.280",
    pages = "3576--3588"
}

@inproceedings{peters-etal-2019-tune,
    title = "To Tune or Not to Tune? Adapting Pretrained Representations to Diverse Tasks",
    author = "Peters, Matthew E.  and
      Ruder, Sebastian  and
      Smith, Noah A.",
    editor = "Augenstein, Isabelle  and
      Gella, Spandana  and
      Ruder, Sebastian  and
      Kann, Katharina  and
      Can, Burcu  and
      Welbl, Johannes  and
      Conneau, Alexis  and
      Ren, Xiang  and
      Rei, Marek",
    booktitle = "Proceedings of the 4th Workshop on Representation Learning for NLP (RepL4NLP-2019)",
    month = aug,
    year = "2019",
    address = "Florence, Italy",
    publisher = "Association for Computational Linguistics",
    url = "https://aclanthology.org/W19-4302/",
    doi = "10.18653/v1/W19-4302",
    pages = "7--14"
}

@article{choenni-shutova-2022-investigating,
    title = "Investigating Language Relationships in Multilingual Sentence Encoders Through the Lens of Linguistic Typology",
    author = "Choenni, Rochelle  and
      Shutova, Ekaterina",
    journal = "Computational Linguistics",
    volume = "48",
    number = "3",
    month = sep,
    year = "2022",
    address = "Cambridge, MA",
    publisher = "MIT Press",
    url = "https://aclanthology.org/2022.cl-3.5/",
    doi = "10.1162/coli_a_00444",
    pages = "635--672"
}

@inproceedings{
wei2021on,
title={On Learning Universal Representations Across Languages},
author={Xiangpeng Wei and Rongxiang Weng and Yue Hu and Luxi Xing and Heng Yu and Weihua Luo},
booktitle={International Conference on Learning Representations},
year={2021},
url={https://openreview.net/forum?id=Uu1Nw-eeTxJ}
}

@inproceedings{wang-etal-2023-gradsim,
    title = "{G}rad{S}im: Gradient-Based Language Grouping for Effective Multilingual Training",
    author = {Wang, Mingyang  and
      Adel, Heike  and
      Lange, Lukas  and
      Str{\"o}tgen, Jannik  and
      Schuetze, Hinrich},
    editor = "Bouamor, Houda  and
      Pino, Juan  and
      Bali, Kalika",
    booktitle = "Proceedings of the 2023 Conference on Empirical Methods in Natural Language Processing",
    month = dec,
    year = "2023",
    address = "Singapore",
    publisher = "Association for Computational Linguistics",
    url = "https://aclanthology.org/2023.emnlp-main.282/",
    doi = "10.18653/v1/2023.emnlp-main.282",
    pages = "4631--4646"
}

@inproceedings{petrov-tokenizers-unfairness,
author = {Petrov, Aleksandar and Malfa, Emanuele La and Torr, Philip H.S. and Bibi, Adel},
title = {Language model tokenizers introduce unfairness between languages},
year = {2023},
publisher = {Curran Associates Inc.},
address = {Red Hook, NY, USA},
booktitle = {Proceedings of the 37th International Conference on Neural Information Processing Systems},
articleno = {1608},
numpages = {28},
location = {New Orleans, LA, USA}
}

@inproceedings{mousi-shared,
    title = "Exploring Alignment in Shared Cross-lingual Spaces",
    author = "Mousi, Basel  and
      Durrani, Nadir  and
      Dalvi, Fahim  and
      Hawasly, Majd  and
      Abdelali, Ahmed",
    editor = "Ku, Lun-Wei  and
      Martins, Andre  and
      Srikumar, Vivek",
    booktitle = "Proceedings of the 62nd Annual Meeting of the Association for Computational Linguistics (Volume 1: Long Papers)",
    month = aug,
    year = "2024",
    address = "Bangkok, Thailand",
    publisher = "Association for Computational Linguistics",
    url = "https://aclanthology.org/2024.acl-long.344/",
    doi = "10.18653/v1/2024.acl-long.344",
    pages = "6326--6348"
}

@inproceedings{mothello,
    title = "m{O}thello: When Do Cross-Lingual Representation Alignment and Cross-Lingual Transfer Emerge in Multilingual Models?",
    author = "Hua, Tianze  and
      Yun, Tian  and
      Pavlick, Ellie",
    editor = "Duh, Kevin  and
      Gomez, Helena  and
      Bethard, Steven",
    booktitle = "Findings of the Association for Computational Linguistics: NAACL 2024",
    month = jun,
    year = "2024",
    address = "Mexico City, Mexico",
    publisher = "Association for Computational Linguistics",
    url = "https://aclanthology.org/2024.findings-naacl.103/",
    doi = "10.18653/v1/2024.findings-naacl.103",
    pages = "1585--1598"
}

@inproceedings{
wu-semantic-hubness,
title={The Semantic Hub Hypothesis: Language Models Share Semantic Representations Across Languages and Modalities},
author={Zhaofeng Wu and Xinyan Velocity Yu and Dani Yogatama and Jiasen Lu and Yoon Kim},
booktitle={International Conference on Learning Representations},
year={2025},
url={https://openreview.net/forum?id=FrFQpAgnGE}
}

@inproceedings{dufter-schutze-2020-identifying,
    title = "Identifying Elements Essential for {BERT}{'}s Multilinguality",
    author = {Dufter, Philipp  and
      Sch{\"u}tze, Hinrich},
    editor = "Webber, Bonnie  and
      Cohn, Trevor  and
      He, Yulan  and
      Liu, Yang",
    booktitle = "Proceedings of the 2020 Conference on Empirical Methods in Natural Language Processing (EMNLP)",
    month = nov,
    year = "2020",
    address = "Online",
    publisher = "Association for Computational Linguistics",
    url = "https://aclanthology.org/2020.emnlp-main.358/",
    doi = "10.18653/v1/2020.emnlp-main.358",
    pages = "4423--4437"
}

@inproceedings{pires-etal-2019-multilingual,
    title = "How Multilingual is Multilingual {BERT}?",
    author = "Pires, Telmo  and
      Schlinger, Eva  and
      Garrette, Dan",
    editor = "Korhonen, Anna  and
      Traum, David  and
      M{\`a}rquez, Llu{\'i}s",
    booktitle = "Proceedings of the 57th Annual Meeting of the Association for Computational Linguistics",
    month = jul,
    year = "2019",
    address = "Florence, Italy",
    publisher = "Association for Computational Linguistics",
    url = "https://aclanthology.org/P19-1493/",
    doi = "10.18653/v1/P19-1493",
    pages = "4996--5001"
}

@inproceedings{
muse-dictionary,
title={Word Translation Without Parallel Data},
author={Guillaume Lample and Alexis Conneau and Marc'Aurelio Ranzato and Ludovic Denoyer and Hervé Jégou},
booktitle={International Conference on Learning Representations},
year={2018},
url={https://openreview.net/forum?id=H196sainb},
}

@inproceedings{bert-paper,
    title = "{BERT}: Pre-training of Deep Bidirectional Transformers for Language Understanding",
    author = "Devlin, Jacob  and
      Chang, Ming-Wei  and
      Lee, Kenton  and
      Toutanova, Kristina",
    editor = "Burstein, Jill  and
      Doran, Christy  and
      Solorio, Thamar",
    booktitle = "Proceedings of the 2019 Conference of the North {A}merican Chapter of the Association for Computational Linguistics: Human Language Technologies, Volume 1 (Long and Short Papers)",
    month = jun,
    year = "2019",
    address = "Minneapolis, Minnesota",
    publisher = "Association for Computational Linguistics",
    url = "https://aclanthology.org/N19-1423/",
    doi = "10.18653/v1/N19-1423",
    pages = "4171--4186"
}

@inproceedings{sib,
    title = "{SIB}-200: A Simple, Inclusive, and Big Evaluation Dataset for Topic Classification in 200+ Languages and Dialects",
    author = "Adelani, David Ifeoluwa  and
      Liu, Hannah  and
      Shen, Xiaoyu  and
      Vassilyev, Nikita  and
      Alabi, Jesujoba O.  and
      Mao, Yanke  and
      Gao, Haonan  and
      Lee, En-Shiun Annie",
    editor = "Graham, Yvette  and
      Purver, Matthew",
    booktitle = "Proceedings of the 18th Conference of the European Chapter of the Association for Computational Linguistics (Volume 1: Long Papers)",
    month = mar,
    year = "2024",
    address = "St. Julian{'}s, Malta",
    publisher = "Association for Computational Linguistics",
    url = "https://aclanthology.org/2024.eacl-long.14/",
    doi = "10.18653/v1/2024.eacl-long.14",
    pages = "226--245"
}

@inproceedings{nguyen-etal-2024-culturax,
    title = "{C}ultura{X}: A Cleaned, Enormous, and Multilingual Dataset for Large Language Models in 167 Languages",
    author = "Nguyen, Thuat  and
      Nguyen, Chien Van  and
      Lai, Viet Dac  and
      Man, Hieu  and
      Ngo, Nghia Trung  and
      Dernoncourt, Franck  and
      Rossi, Ryan A.  and
      Nguyen, Thien Huu",
    editor = "Calzolari, Nicoletta  and
      Kan, Min-Yen  and
      Hoste, Veronique  and
      Lenci, Alessandro  and
      Sakti, Sakriani  and
      Xue, Nianwen",
    booktitle = "Proceedings of the 2024 Joint International Conference on Computational Linguistics, Language Resources and Evaluation (LREC-COLING 2024)",
    month = may,
    year = "2024",
    address = "Torino, Italia",
    publisher = "ELRA and ICCL",
    url = "https://aclanthology.org/2024.lrec-main.377",
    pages = "4226--4237",
}

@inproceedings{penedo2024finewebdatasetsdecantingweb,
author = {Penedo, Guilherme and Kydl\'{\i}\v{c}ek, Hynek and Allal, Loubna Ben and Lozhkov, Anton and Mitchell, Margaret and Raffel, Colin and Von Werra, Leandro and Wolf, Thomas},
title = {The FineWeb datasets: decanting the web for the finest text data at scale},
year = {2025},
isbn = {9798331314385},
publisher = {Curran Associates Inc.},
address = {Red Hook, NY, USA},
booktitle = {Proceedings of the 38th International Conference on Neural Information Processing Systems},
articleno = {970},
numpages = {39},
location = {Vancouver, BC, Canada}
}

@inproceedings{curse-multilinguality,
    title = "Unsupervised Cross-lingual Representation Learning at Scale",
    author = "Conneau, Alexis  and
      Khandelwal, Kartikay  and
      Goyal, Naman  and
      Chaudhary, Vishrav  and
      Wenzek, Guillaume  and
      Guzm{\'a}n, Francisco  and
      Grave, Edouard  and
      Ott, Myle  and
      Zettlemoyer, Luke  and
      Stoyanov, Veselin",
    editor = "Jurafsky, Dan  and
      Chai, Joyce  and
      Schluter, Natalie  and
      Tetreault, Joel",
    booktitle = "Proceedings of the 58th Annual Meeting of the Association for Computational Linguistics",
    month = jul,
    year = "2020",
    address = "Online",
    publisher = "Association for Computational Linguistics",
    url = "https://aclanthology.org/2020.acl-main.747/",
    doi = "10.18653/v1/2020.acl-main.747",
    pages = "8440--8451"
}

@inproceedings{ustun-etal-2024-aya,
    title = "Aya Model: An Instruction Finetuned Open-Access Multilingual Language Model",
    author = {{\"U}st{\"u}n, Ahmet  and
      Aryabumi, Viraat  and
      Yong, Zheng  and
      Ko, Wei-Yin  and
      D{'}souza, Daniel  and
      Onilude, Gbemileke  and
      Bhandari, Neel  and
      Singh, Shivalika  and
      Ooi, Hui-Lee  and
      Kayid, Amr  and
      Vargus, Freddie  and
      Blunsom, Phil  and
      Longpre, Shayne  and
      Muennighoff, Niklas  and
      Fadaee, Marzieh  and
      Kreutzer, Julia  and
      Hooker, Sara},
    editor = "Ku, Lun-Wei  and
      Martins, Andre  and
      Srikumar, Vivek",
    booktitle = "Proceedings of the 62nd Annual Meeting of the Association for Computational Linguistics (Volume 1: Long Papers)",
    month = aug,
    year = "2024",
    address = "Bangkok, Thailand",
    publisher = "Association for Computational Linguistics",
    url = "https://aclanthology.org/2024.acl-long.845/",
    doi = "10.18653/v1/2024.acl-long.845",
    pages = "15894--15939"
}

@misc{apertus,
      title={Apertus: Democratizing Open and Compliant LLMs for Global Language Environments}, 
      author={Alejandro Hernández-Cano and Alexander Hägele and Allen Hao Huang and Angelika Romanou and Antoni-Joan Solergibert and Barna Pasztor and Bettina Messmer and Dhia Garbaya and Eduard Frank Ďurech and Ido Hakimi and Juan García Giraldo and Mete Ismayilzada and Negar Foroutan and Skander Moalla and Tiancheng Chen and Vinko Sabolčec and Yixuan Xu and Michael Aerni and Badr AlKhamissi and Ines Altemir Marinas and Mohammad Hossein Amani and Matin Ansaripour and Ilia Badanin and Harold Benoit and Emanuela Boros and Nicholas Browning and Fabian Bösch and Maximilian Böther and Niklas Canova and Camille Challier and Clement Charmillot and Jonathan Coles and Jan Deriu and Arnout Devos and Lukas Drescher and Daniil Dzenhaliou and Maud Ehrmann and Dongyang Fan and Simin Fan and Silin Gao and Miguel Gila and María Grandury and Diba Hashemi and Alexander Hoyle and Jiaming Jiang and Mark Klein and Andrei Kucharavy and Anastasiia Kucherenko and Frederike Lübeck and Roman Machacek and Theofilos Manitaras and Andreas Marfurt and Kyle Matoba and Simon Matrenok and Henrique Mendoncça and Fawzi Roberto Mohamed and Syrielle Montariol and Luca Mouchel and Sven Najem-Meyer and Jingwei Ni and Gennaro Oliva and Matteo Pagliardini and Elia Palme and Andrei Panferov and Léo Paoletti and Marco Passerini and Ivan Pavlov and Auguste Poiroux and Kaustubh Ponkshe and Nathan Ranchin and Javi Rando and Mathieu Sauser and Jakhongir Saydaliev and Muhammad Ali Sayfiddinov and Marian Schneider and Stefano Schuppli and Marco Scialanga and Andrei Semenov and Kumar Shridhar and Raghav Singhal and Anna Sotnikova and Alexander Sternfeld and Ayush Kumar Tarun and Paul Teiletche and Jannis Vamvas and Xiaozhe Yao and Hao Zhao Alexander Ilic and Ana Klimovic and Andreas Krause and Caglar Gulcehre and David Rosenthal and Elliott Ash and Florian Tramèr and Joost VandeVondele and Livio Veraldi and Martin Rajman and Thomas Schulthess and Torsten Hoefler and Antoine Bosselut and Martin Jaggi and Imanol Schlag},
      year={2025},
      eprint={2509.14233},
      archivePrefix={arXiv},
      primaryClass={cs.CL},
      url={https://arxiv.org/abs/2509.14233}, 
}

@inproceedings{liu-etal-2025-transliterations,
    title = "How Transliterations Improve Crosslingual Alignment",
    author = {Liu, Yihong  and
      Wang, Mingyang  and
      Kargaran, Amir Hossein  and
      ImaniGooghari, Ayyoob  and
      Xhelili, Orgest  and
      Ye, Haotian  and
      Ma, Chunlan  and
      Yvon, Fran{\c{c}}ois  and
      Sch{\"u}tze, Hinrich},
    editor = "Rambow, Owen  and
      Wanner, Leo  and
      Apidianaki, Marianna  and
      Al-Khalifa, Hend  and
      Eugenio, Barbara Di  and
      Schockaert, Steven",
    booktitle = "Proceedings of the 31st International Conference on Computational Linguistics",
    month = jan,
    year = "2025",
    address = "Abu Dhabi, UAE",
    publisher = "Association for Computational Linguistics",
    url = "https://aclanthology.org/2025.coling-main.165/",
    pages = "2417--2433"
}

\appendix
\section{Strong and Weak Cross-Lingual Alignment}
\label{app_a_alignment_def}
We formalize \textit{strong} and \textit{weak} cross-lingual alignment as follows. Let $\mathcal{S}=\{L_1,\dots,L_n\}$ be the set of languages and $\mathcal{X}_L$ the input space for language $L$. An encoder
\begin{equation}
\label{eq:encoder}
f_\theta:\bigcup_{L\in\mathcal{S}}\mathcal{X}_L \to \mathbb{R}^d
\end{equation}
maps inputs into a shared representation space, and $\operatorname{sim}:\mathbb{R}^d\times\mathbb{R}^d\to\mathbb{R}$ measures embedding similarity. We define a semantic equivalence relation $x\sim x'$ for pairs expressing the same meaning and polarity; non-equivalent $x\not\sim x'$ includes unrelated or opposite content. Let $\mathcal{D}_{\mathrm{pos}}$ and $\mathcal{D}_{\mathrm{neg}}$ denote distributions over equivalent and non-equivalent pairs.

\paragraph{Strong Alignment.}
The encoder $f_\theta$ satisfies \textit{$(\delta,\epsilon)$-strong alignment} between $L_a$ and $L_b$ if, with probability at least $1-\epsilon$ over $(x,x')\!\sim\!\mathcal{D}_{\mathrm{pos}}$ and $(x,x'')\!\sim\!\mathcal{D}_{\mathrm{neg}}$,
\begin{equation}
\label{eq:probabilistic}
\operatorname{sim}\!\left(f_\theta(x),f_\theta(x')\right)
\ge \operatorname{sim}\!\left(f_\theta(x),f_\theta(x'')\right)+\delta.
\end{equation}
This enforces that equivalent pairs remain more similar than non-equivalent ones by at least a margin $\delta$.

\paragraph{Weak Alignment.}
To relax the binary equivalence assumption, we introduce a graded semantic similarity function
\begin{equation}
\label{eq:sigma}
\sigma:\Big(\!\bigcup_{L}\mathcal{X}_L\!\Big)\times\Big(\!\bigcup_{L}\mathcal{X}_L\!\Big)\to[-1,1],
\end{equation}
with $\sigma(x,x')=1$ for perfect translations and values near $-1$ for antonyms. We instantiate $\operatorname{sim}$ using cosine similarity:
\begin{equation}
\label{eq:cosine}
\operatorname{sim}(u,v)=\frac{u\cdot v}{\|u\|\|v\|}.
\end{equation}
For a distribution $\mathcal{D}$ over arbitrary cross-lingual pairs, $f_\theta$ exhibits \textit{$\gamma$-weak alignment} if
\begin{equation}
\label{eq:weak_alignment}
\mathrm{Cov}_{(x,x')\sim\mathcal{D}}\!\left[\sigma(x,x'),\operatorname{sim}(f_\theta(x),f_\theta(x'))\right]\ge\gamma,
\end{equation}
or equivalently,
\begin{equation}
\label{eq:weak_alignment_corr}
\mathrm{Corr}_{(x,x')\sim\mathcal{D}}\!\left[\sigma(x,x'),\operatorname{sim}(f_\theta(x),f_\theta(x'))\right]\ge\rho,
\end{equation}
with $\rho=\gamma/\sqrt{\mathrm{Var}[\sigma]\mathrm{Var}[\operatorname{sim}]}$. Weak alignment thus requires embedding similarity to increase with semantic relatedness without assuming strict translation equivalence.

\section{Hyperparameters and Computational Resources}
\label{app_b_hyperparams}

\begin{table}[h]
\centering
\resizebox{0.35\textwidth}{!}{
\begin{tabular}{l c}
\toprule
\textbf{Parameter} & \textbf{Value} \\
\midrule
Learning rate & $1e-5$ \\
Batch size & $16$ \\
Max sequence length & $128$ \\
Word negatives & $512$ \\
Temperature & $0.1$ \\
Number of epochs & $1$ \\ 
Number of steps & $\approx 6,000$ \\
Number of sentences\textsuperscript{*} & $\approx 100,000$ \\
Seed & $42$ \\
\bottomrule
\end{tabular}}
\caption{Hyperparameters used during alignment enhancement (Phase I). \textsuperscript{*} each language pair uses 100,000 sentences, except English-Kazakh which only has 80,000 sentences available.}
\label{tab:alignment_params}
\end{table}

\begin{table}[h!]
  \centering
  \begin{tabular}{lcc}
    \toprule
    \textbf{Parameter} & \textbf{Value}\\
    \midrule
    Batch size & 16 \\
    Backbone learning rate (full model) & $1e-4$ \\
    Classifier learning rate (full model) & $1e-3$ \\
    Learning rate (linear layer only) & $5e-3$ \\
    Number of epochs & 15 \\
    Weight decay & 0.01 \\
    Max sequence length & 256 \\
    Seeds & 15, 33, 42 \\
    \bottomrule
  \end{tabular}
  \caption{Hyperparameters used for full model and linear layer only fine-tuning (Phase II) for POS-Tagging.}
  \label{tab:pos-params}
\end{table}

\begin{table}[h!]
  \centering
  \begin{tabular}{lcc}
    \toprule
    \textbf{Parameter} & \textbf{Value} \\
    \midrule
    Batch size & 16 \\
    Learning rate (full model) & $2e-5$ \\
    Learning rate (linear layer only) & $3e-3$  \\
    Number of epochs & 15 \\
    Max sequence length & 256 \\
    Seeds & 15, 33, 42 \\
    \bottomrule
  \end{tabular}
  \caption{Hyperparameters used for full model and linear layer only fine-tuning (Phase II) for Sentence Classification.}
  \label{tab:sib-params}
\end{table}

In Phase I (alignment fine-tuning), we use a single V100 NVIDIA GPUxt to tune each variant of \xlmr for each language pair (7 loss configurations per alignment pair, a total of 28 models). Total amount of GPUh spent on alignment fine-tuning across models is $\approx 275$, with individual model tuning time varying from 5 to 22 GPUh. \\
In Phase II for \textbf{POS Tagging fine-tuning}, we use a single V100 NVIDIA GPU to fine-tune each variant of aligned \xlmr, i.e. \xlmrdeen, \xlmrenkk, \xlmrjakk, \xlmrkkko for \textsc{POS Tagging} as well as the baseline (\xlmr) itself. We fine-tune the whole model as well as the linear layer only.  Total amount of GPUh spent on \textsc{POS Tagging} fine-tuning across models is $\approx 30$. \\
In Phase II for \textbf{Sentence Classification fine-tuning} we use 1xV100 GPU to fine-tune each variant of aligned \xlmr, i.e. \xlmrdeen, \xlmrenkk, \xlmrjakk, \xlmrkkko for \textsc{Sentence Classification} as well as the baseline (\xlmr) itself. We fine-tune the whole model as well as the linear layer only. Total amount of GPUh spent on \textsc{Sentence Classification} fine-tuning across models is $\approx 8$.

\newpage
\begin{figure*}
    \centering
    \includegraphics[width=1\linewidth]{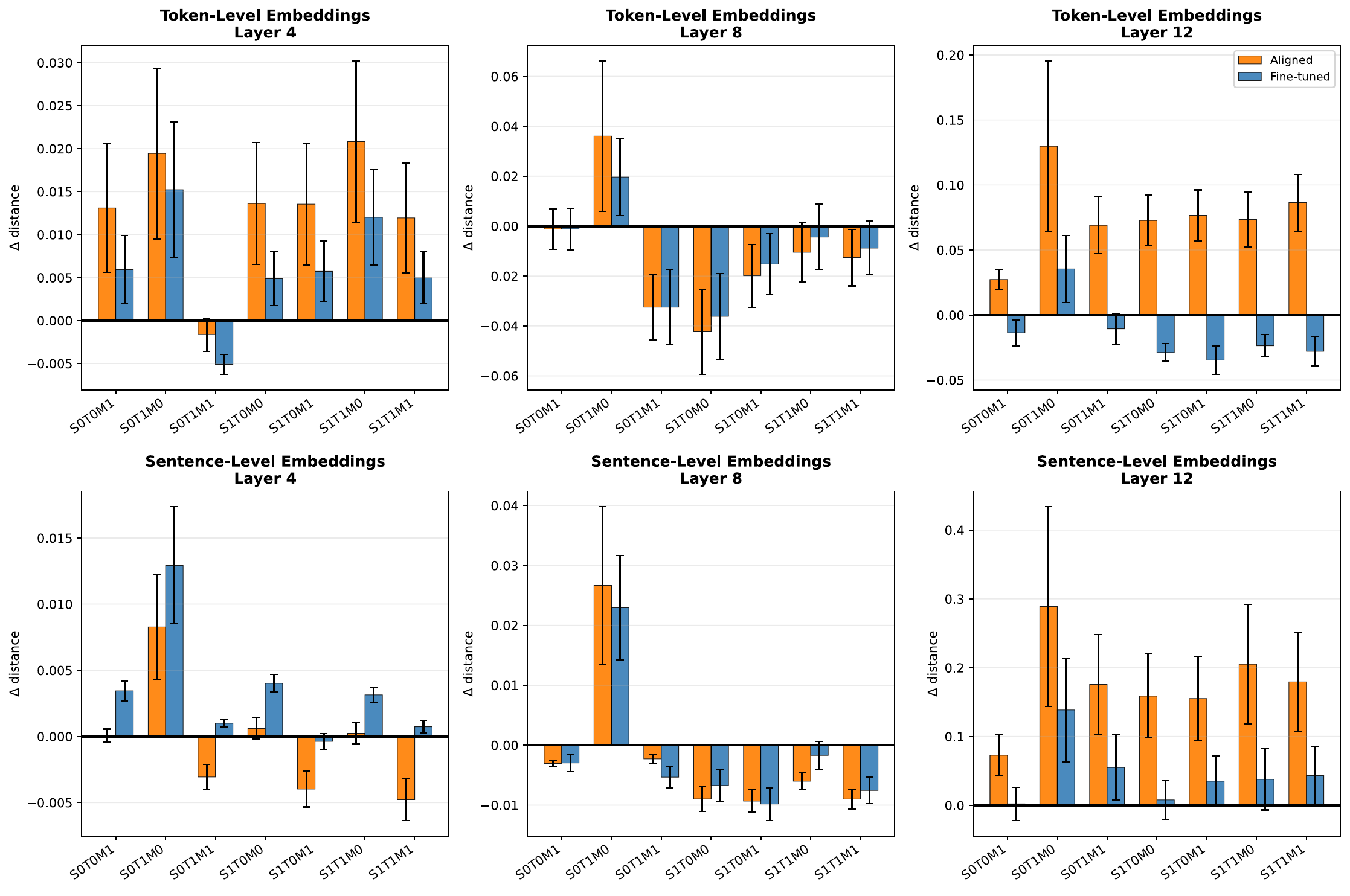}
    \caption{$\Delta$ in cosine distance from \xlmr for embeddings
    averaged across models and languages and grouped by loss configuration across 4\textsuperscript{th}, 8\textsuperscript{th}, and 12\textsuperscript{th} layers.}
    \label{fig:cos-dists}
\end{figure*}

\begin{table*}[hbtp!]
\section{Evaluation Details}
\label{app_c_recall_acc}
\centering
\small
\resizebox{\textwidth}{!}{%
\begin{tabular}{lccccccc}
\toprule
Lang Pair & S0T0M1 & S0T1M0 & S0T1M1 & S1T0M0 & S1T0M1 & S1T1M0 & S1T1M1 \\
\midrule
\multicolumn{8}{c}{\textbf{Group 1: Aligned Languages With English}} \\
\midrule
deu-eng & $\phantom{-}0.00 \pm 0.00$ & \textcolor{red}{$-0.09 \pm 0.05$} &  $\phantom{-}0.00 \pm 0.00$ & $\phantom{-}0.00 \pm 0.00$ & $\phantom{-}0.00 \pm 0.00$ & $\phantom{-}0.00 \pm 0.00$ & $\phantom{-}0.00 \pm 0.00$ \\
kaz-eng & \textcolor{green}{$\phantom{-}0.04 \pm 0.01$} & \textcolor{red}{$-0.67 \pm 0.22$} & $-0.03 \pm 0.07$ & \textcolor{green}{$\phantom{-}0.06 \pm 0.02$} & \textcolor{green}{$\phantom{-}0.05 \pm 0.02$} & $\phantom{-}0.03 \pm 0.05$ & $\phantom{-}0.04 \pm 0.04$ \\
kor-eng & \textcolor{green}{$\phantom{-}0.02 \pm 0.01$} & \textcolor{red}{$-0.58 \pm 0.22$} & $-0.01 \pm 0.03$ & \textcolor{green}{$\phantom{-}0.02 \pm 0.01$} & $\phantom{-}0.01 \pm 0.02$ & $-0.01 \pm 0.04$ & $-0.02 \pm 0.04$ \\
jpn-eng & $\phantom{-}0.01 \pm 0.02$ & \textcolor{red}{$-0.49 \pm 0.09$} & \textcolor{red}{$-0.03 \pm 0.02$} & \textcolor{green}{$\phantom{-}0.02 \pm 0.01$} & $\phantom{-}0.01 \pm 0.01$ & \textcolor{green}{$\phantom{-}0.02 \pm 0.01$} & $\phantom{-}0.01 \pm 0.01$ \\

\hline
\addlinespace
\multicolumn{8}{c}{\textbf{Group 2: Aligned Languages Without English}} \\
\midrule
jpn-kaz & \textcolor{green}{$\phantom{-}0.02 \pm 0.01$} & \textcolor{red}{$-0.51 \pm 0.13$} & $-0.02 \pm 0.03$ & \textcolor{green}{$\phantom{-}0.04 \pm 0.02$} & \textcolor{green}{$\phantom{-}0.05 \pm 0.02$} & \textcolor{green}{$\phantom{-}0.03 \pm 0.02$} & $\phantom{-}0.03 \pm 0.03$ \\
kor-kaz & \textcolor{green}{$\phantom{-}0.03 \pm 0.01$} & \textcolor{red}{$-0.41 \pm 0.13$} & $\phantom{-}0.00 \pm 0.03$ & \textcolor{green}{$\phantom{-}0.04 \pm 0.01$} & \textcolor{green}{$\phantom{-}0.04 \pm 0.01$} & \textcolor{green}{$\phantom{-}0.03 \pm 0.02$} & $\phantom{-}0.02 \pm 0.02$ \\

\hline
\addlinespace
\multicolumn{8}{c}{\textbf{Group 3: Other Languages With English}} \\
\midrule
hin-eng & \textcolor{green}{$\phantom{-}0.03 \pm 0.01$} & \textcolor{red}{$-0.48 \pm 0.15$} & $-0.03 \pm 0.03$ & \textcolor{green}{$\phantom{-}0.03 \pm 0.01$} & \textcolor{green}{$\phantom{-}0.02 \pm 0.01$} & $\phantom{-}0.01 \pm 0.02$ & $\phantom{-}0.00 \pm 0.03$ \\
zho-eng & \textcolor{green}{$\phantom{-}0.02 \pm 0.00$} & \textcolor{red}{$-0.32 \pm 0.08$} & $\phantom{-}0.00 \pm 0.01$ & \textcolor{green}{$\phantom{-}0.02 \pm 0.01$} & \textcolor{green}{$\phantom{-}0.02 \pm 0.00$} & \textcolor{green}{$\phantom{-}0.02 \pm 0.01$} & \textcolor{green}{$\phantom{-}0.02 \pm 0.01$} \\
spa-eng & $\phantom{-}0.00 \pm 0.00$ & \textcolor{red}{$-0.06 \pm 0.03$} & $\phantom{-}0.00 \pm 0.00$ & $\phantom{-}0.00 \pm 0.00$ & $\phantom{-}0.00 \pm 0.00$ & $\phantom{-}0.00 \pm 0.00$ & $\phantom{-}0.00 \pm 0.00$ \\
rus-eng & $\phantom{-}0.00 \pm 0.00$ & \textcolor{red}{$-0.29 \pm 0.15$} & $\phantom{-}0.00 \pm 0.01$ & $\phantom{-}0.00 \pm 0.00$ & \textcolor{green}{$\phantom{-}0.01 \pm 0.00$} & $\phantom{-}0.00 \pm 0.00$ & $\phantom{-}0.00 \pm 0.00$ \\

\hline
\addlinespace
\multicolumn{8}{c}{\textbf{Group 4: Languages of the Same Script, Non-Aligned}} \\
\midrule
lit-spa & \textcolor{green}{$\phantom{-}0.02 \pm 0.00$} & \textcolor{red}{$-0.30 \pm 0.12$} & $\phantom{-}0.01 \pm 0.02$ & \textcolor{green}{$\phantom{-}0.04 \pm 0.01$} & \textcolor{green}{$\phantom{-}0.04 \pm 0.01$} & \textcolor{green}{$\phantom{-}0.04 \pm 0.01$} & \textcolor{green}{$\phantom{-}0.03 \pm 0.01$} \\
hin-kas & $\phantom{-}0.00 \pm 0.01$ & \textcolor{red}{$-0.02 \pm 0.01$} & $-0.02 \pm 0.02$ & \textcolor{green}{$\phantom{-}0.06 \pm 0.02$} & \textcolor{green}{$\phantom{-}0.03 \pm 0.01$} & \textcolor{green}{$\phantom{-}0.06 \pm 0.04$} & \textcolor{green}{$\phantom{-}0.04 \pm 0.01$} \\
spa-fra & $\phantom{-}0.00 \pm 0.00$ & \textcolor{red}{$-0.06 \pm 0.03$} & $\phantom{-}0.00 \pm 0.00$ & $\phantom{-}0.00 \pm 0.00$ & $\phantom{-}0.00 \pm 0.00$ & $\phantom{-}0.00 \pm 0.00$ & $\phantom{-}0.00 \pm 0.00$ \\
rus-bul & $\phantom{-}0.00 \pm 0.00$ & \textcolor{red}{$-0.05 \pm 0.01$} & $\phantom{-}0.00 \pm 0.00$ & $\phantom{-}0.00 \pm 0.00$ & $\phantom{-}0.00 \pm 0.00$ & $\phantom{-}0.00 \pm 0.00$ & $\phantom{-}0.00 \pm 0.00$ \\

\hline
\addlinespace
\multicolumn{8}{c}{\textbf{Group 5: Languages of the Same Script, Aligned}} \\
\midrule
kaz-bul & \textcolor{green}{$\phantom{-}0.06 \pm 0.00$} & \textcolor{red}{$-0.46 \pm 0.17$} & $\phantom{-}0.02 \pm 0.04$ & \textcolor{green}{$\phantom{-}0.08 \pm 0.01$} & \textcolor{green}{$\phantom{-}0.08 \pm 0.01$} & \textcolor{green}{$\phantom{-}0.06 \pm 0.05$} & \textcolor{green}{$\phantom{-}0.06 \pm 0.02$} \\
kaz-rus & \textcolor{green}{$\phantom{-}0.05 \pm 0.01$} & \textcolor{red}{$-0.45 \pm 0.18$} & $\phantom{-}0.03 \pm 0.03$ & \textcolor{green}{$\phantom{-}0.08 \pm 0.02$} & \textcolor{green}{$\phantom{-}0.08 \pm 0.01$} & \textcolor{green}{$\phantom{-}0.06 \pm 0.05$} & \textcolor{green}{$\phantom{-}0.07 \pm 0.01$} \\
deu-spa & \textcolor{green}{$\phantom{-}0.01 \pm 0.00$} & \textcolor{red}{$-0.17 \pm 0.07$} & $\phantom{-}0.00 \pm 0.00$ & $\phantom{-}0.01 \pm 0.01$ & \textcolor{green}{$\phantom{-}0.01 \pm 0.00$} & \textcolor{green}{$\phantom{-}0.01 \pm 0.00$} & \textcolor{green}{$\phantom{-}0.01 \pm 0.00$} \\
deu-fra & \textcolor{green}{$\phantom{-}0.01 \pm 0.00$} & \textcolor{red}{$-0.13 \pm 0.07$} & $\phantom{-}0.00 \pm 0.00$ & $\phantom{-}0.01 \pm 0.01$ & \textcolor{green}{$\phantom{-}0.01 \pm 0.00$} & $\phantom{-}0.01 \pm 0.01$ & \textcolor{green}{$\phantom{-}0.01 \pm 0.00$} \\

\hline
\addlinespace
\multicolumn{8}{c}{\textbf{Group 6: Languages of Different Scripts, Non-Aligned}} \\
\midrule
spa-rus & $\phantom{-}0.00 \pm 0.00$ & \textcolor{red}{$-0.15 \pm 0.05$} & $\phantom{-}0.00 \pm 0.00$ & \textcolor{green}{$\phantom{-}0.01 \pm 0.00$} & \textcolor{green}{$\phantom{-}0.01 \pm 0.00$} & $\phantom{-}0.00 \pm 0.00$ & $\phantom{-}0.00 \pm 0.00$ \\
spa-kas & $\phantom{-}0.01 \pm 0.01$ & $-0.01 \pm 0.01$ & $-0.01 \pm 0.01$ & \textcolor{green}{$\phantom{-}0.05 \pm 0.01$} & \textcolor{green}{$\phantom{-}0.03 \pm 0.01$} & \textcolor{green}{$\phantom{-}0.05 \pm 0.02$} & \textcolor{green}{$\phantom{-}0.03 \pm 0.01$} \\
heb-fra & \textcolor{green}{$\phantom{-}0.03 \pm 0.00$} & \textcolor{red}{$-0.36 \pm 0.10$} & $\phantom{-}0.01 \pm 0.02$ & \textcolor{green}{$\phantom{-}0.04 \pm 0.01$} & \textcolor{green}{$\phantom{-}0.04 \pm 0.01$} & \textcolor{green}{$\phantom{-}0.04 \pm 0.01$} & \textcolor{green}{$\phantom{-}0.03 \pm 0.01$} \\
hin-zho & \textcolor{green}{$\phantom{-}0.03 \pm 0.02$} & \textcolor{red}{$-0.41 \pm 0.05$} & $-0.02 \pm 0.05$ & $\phantom{-}0.03 \pm 0.03$ & \textcolor{green}{$\phantom{-}0.04 \pm 0.03$} & $\phantom{-}0.02 \pm 0.04$ & $\phantom{-}0.02 \pm 0.03$\\
\hline
\addlinespace
\multicolumn{8}{c}{\textbf{Group 7: Languages of Different Scripts, Aligned}} \\
\midrule
deu-heb & \textcolor{green}{$\phantom{-}0.01 \pm 0.00$} & \textcolor{red}{$-0.26 \pm 0.09$} & $\phantom{-}0.00 \pm 0.00$ & $\phantom{-}0.00 \pm 0.00$ & \textcolor{green}{$\phantom{-}0.01 \pm 0.00$} & $\phantom{-}0.00 \pm 0.00$ & $\phantom{-}0.00 \pm 0.01$ \\
kaz-hin & \textcolor{green}{$\phantom{-}0.06 \pm 0.01$} & \textcolor{red}{$-0.42 \pm 0.16$} & $\phantom{-}0.02 \pm 0.05$ & \textcolor{green}{$\phantom{-}0.09 \pm 0.01$} & \textcolor{green}{$\phantom{-}0.09 \pm 0.01$} & \textcolor{green}{$\phantom{-}0.09 \pm 0.01$} & \textcolor{green}{$\phantom{-}0.08 \pm 0.01$} \\
jpn-rus & \textcolor{green}{$\phantom{-}0.03 \pm 0.01$} & \textcolor{red}{$-0.31 \pm 0.04$} & $\phantom{-}0.00 \pm 0.02$ & \textcolor{green}{$\phantom{-}0.05 \pm 0.01$} & \textcolor{green}{$\phantom{-}0.05 \pm 0.01$} & \textcolor{green}{$\phantom{-}0.04 \pm 0.02$} & \textcolor{green}{$\phantom{-}0.04 \pm 0.01$} \\
kor-fra & \textcolor{green}{$\phantom{-}0.06 \pm 0.01$} & \textcolor{red}{$-0.53 \pm 0.17$} & $\phantom{-}0.01 \pm 0.05$ & \textcolor{green}{$\phantom{-}0.07 \pm 0.02$} & \textcolor{green}{$\phantom{-}0.08 \pm 0.01$} & $\phantom{-}0.05 \pm 0.05$ & $\phantom{-}0.04 \pm 0.04$ \\
\bottomrule
\end{tabular}
}
\caption{R@5 $\Delta$ from baseline (\xlmr), averaged over all aligned models and grouped by loss configuration. Color coding indicates the averaged performance change magnitude: \textcolor{red}{red} for degradations, \textcolor{green}{green} for gains (adjusted for standard deviation).}
\label{tab:delta-r5}
\end{table*}

\begin{table*}[hbtp]
\centering
\small
\resizebox{\textwidth}{!}{%
\begin{tabular}{lccccccc}
\toprule
Lang Pair & S0T0M1 & S0T1M0 & S0T1M1 & S1T0M0 & S1T0M1 & S1T1M0 & S1T1M1 \\
\midrule
\multicolumn{8}{c}{\textbf{Group 1: Aligned Languages With English}} \\
\midrule
deu-eng & $\phantom{-}0.00 \pm 0.00$ & \textcolor{red}{$-0.07 \pm 0.04$} & $\phantom{-}0.00 \pm 0.00$ & $\phantom{-}0.00 \pm 0.00$ & $\phantom{-}0.00 \pm 0.00$ & $\phantom{-}0.00 \pm 0.00$ & $\phantom{-}0.00 \pm 0.00$ \\
kaz-eng & \textcolor{green}{$\phantom{-}0.03 \pm 0.01$} & \textcolor{red}{$-0.64 \pm 0.23$} & $-0.03 \pm 0.05$ & \textcolor{green}{$\phantom{-}0.04 \pm 0.02$} & \textcolor{green}{$\phantom{-}0.04 \pm 0.02$} & $\phantom{-}0.02 \pm 0.04$ & $\phantom{-}0.02 \pm 0.02$ \\
kor-eng & \textcolor{green}{$\phantom{-}0.01 \pm 0.00$} & \textcolor{red}{$-0.52 \pm 0.23$} & $-0.01 \pm 0.02$ & $\phantom{-}0.01 \pm 0.01$ & $\phantom{-}0.00 \pm 0.01$ & $-0.01 \pm 0.03$ & $-0.01 \pm 0.03$ \\
jpn-eng & $\phantom{-}0.00 \pm 0.01$ & \textcolor{red}{$-0.42 \pm 0.08$} & \textcolor{red}{$-0.02 \pm 0.01$} & $\phantom{-}0.01 \pm 0.01$ & $\phantom{-}0.01 \pm 0.01$ & $\phantom{-}0.01 \pm 0.01$ & $\phantom{-}0.00 \pm 0.01$ \\

\hline
\addlinespace
\multicolumn{8}{c}{\textbf{Group 2: Aligned Languages Without English}} \\
\midrule
jpn-kaz & \textcolor{green}{$\phantom{-}0.02 \pm 0.01$} & \textcolor{red}{$-0.46 \pm 0.13$} & $-0.01 \pm 0.03$ & \textcolor{green}{$\phantom{-}0.03 \pm 0.01$} & \textcolor{green}{$\phantom{-}0.04 \pm 0.01$} & $\phantom{-}0.02 \pm 0.02$ & \textcolor{green}{$\phantom{-}0.02 \pm 0.01$} \\
kor-kaz & \textcolor{green}{$\phantom{-}0.03 \pm 0.01$} & \textcolor{red}{$-0.34 \pm 0.11$} & $\phantom{-}0.01 \pm 0.02$ & \textcolor{green}{$\phantom{-}0.03 \pm 0.01$} & \textcolor{green}{$\phantom{-}0.03 \pm 0.01$} & \textcolor{green}{$\phantom{-}0.03 \pm 0.01$} & \textcolor{green}{$\phantom{-}0.02 \pm 0.01$} \\

\hline
\addlinespace
\multicolumn{8}{c}{\textbf{Group 3: Other Languages With English}} \\
\midrule
hin-eng & \textcolor{green}{$\phantom{-}0.02 \pm 0.00$} & \textcolor{red}{$-0.43 \pm 0.16$} & $-0.02 \pm 0.02$ & \textcolor{green}{$\phantom{-}0.02 \pm 0.01$} & \textcolor{green}{$\phantom{-}0.02 \pm 0.01$} & $\phantom{-}0.01 \pm 0.01$ & $\phantom{-}0.00 \pm 0.02$ \\
zho-eng & \textcolor{green}{$\phantom{-}0.01 \pm 0.00$} & \textcolor{red}{$-0.26 \pm 0.08$} & $\phantom{-}0.00 \pm 0.01$ & \textcolor{green}{$\phantom{-}0.02 \pm 0.01$} & \textcolor{green}{$\phantom{-}0.01 \pm 0.00$} & \textcolor{green}{$\phantom{-}0.01 \pm 0.00$} & \textcolor{green}{$\phantom{-}0.01 \pm 0.00$} \\
spa-eng & $\phantom{-}0.00 \pm 0.00$ & \textcolor{red}{$-0.04 \pm 0.02$} & $\phantom{-}0.00 \pm 0.00$ & $\phantom{-}0.00 \pm 0.00$ & $\phantom{-}0.00 \pm 0.00$ & $\phantom{-}0.00 \pm 0.00$ & $\phantom{-}0.00 \pm 0.00$ \\
rus-eng & $\phantom{-}0.00 \pm 0.00$ & \textcolor{red}{$-0.23 \pm 0.13$} & $\phantom{-}0.00 \pm 0.01$ & $\phantom{-}0.00 \pm 0.00$ & $\phantom{-}0.00 \pm 0.00$ & $\phantom{-}0.00 \pm 0.00$ & $\phantom{-}0.00 \pm 0.00$ \\

\hline
\addlinespace
\multicolumn{8}{c}{\textbf{Group 4: Languages of the Same Script, Non-Aligned}} \\
\midrule
lit-spa & \textcolor{green}{$\phantom{-}0.02 \pm 0.00$} & \textcolor{red}{$-0.25 \pm 0.10$} & $\phantom{-}0.00 \pm 0.01$ & \textcolor{green}{$\phantom{-}0.03 \pm 0.01$} & \textcolor{green}{$\phantom{-}0.03 \pm 0.00$} & \textcolor{green}{$\phantom{-}0.03 \pm 0.00$} & \textcolor{green}{$\phantom{-}0.02 \pm 0.01$} \\
hin-kas & $\phantom{-}0.00 \pm 0.01$ & \textcolor{red}{$-0.04 \pm 0.01$} & $-0.03 \pm 0.03$ & \textcolor{green}{$\phantom{-}0.07 \pm 0.03$} & \textcolor{green}{$\phantom{-}0.04 \pm 0.02$} & \textcolor{green}{$\phantom{-}0.07 \pm 0.05$} & \textcolor{green}{$\phantom{-}0.04 \pm 0.01$} \\
spa-fra & $\phantom{-}0.00 \pm 0.00$ & \textcolor{red}{$-0.04 \pm 0.02$} & $\phantom{-}0.00 \pm 0.00$ & $\phantom{-}0.00 \pm 0.00$ & $\phantom{-}0.00 \pm 0.00$ & $\phantom{-}0.00 \pm 0.00$ & $\phantom{-}0.00 \pm 0.00$ \\
rus-bul & $\phantom{-}0.00 \pm 0.00$ & \textcolor{red}{$-0.03 \pm 0.01$} & $\phantom{-}0.00 \pm 0.00$ & $\phantom{-}0.00 \pm 0.00$ & $\phantom{-}0.00 \pm 0.00$ & $\phantom{-}0.00 \pm 0.00$ & $\phantom{-}0.00 \pm 0.00$ \\

\hline
\addlinespace
\multicolumn{8}{c}{\textbf{Group 5: Languages of the Same Script, Aligned}} \\
\midrule
kaz-bul & \textcolor{green}{$\phantom{-}0.05 \pm 0.01$} & \textcolor{red}{$-0.42 \pm 0.19$} & $\phantom{-}0.01 \pm 0.03$ & \textcolor{green}{$\phantom{-}0.06 \pm 0.01$} & \textcolor{green}{$\phantom{-}0.06 \pm 0.01$} & $\phantom{-}0.04 \pm 0.04$ & \textcolor{green}{$\phantom{-}0.05 \pm 0.01$} \\
kaz-rus & \textcolor{green}{$\phantom{-}0.04 \pm 0.01$} & \textcolor{red}{$-0.42 \pm 0.19$} & $\phantom{-}0.01 \pm 0.02$ & \textcolor{green}{$\phantom{-}0.06 \pm 0.01$} & \textcolor{green}{$\phantom{-}0.06 \pm 0.01$} & $\phantom{-}0.04 \pm 0.04$ & \textcolor{green}{$\phantom{-}0.05 \pm 0.01$} \\
deu-spa & $\phantom{-}0.00 \pm 0.00$ & \textcolor{red}{$-0.12 \pm 0.06$} & $\phantom{-}0.00 \pm 0.00$ & \textcolor{green}{$\phantom{-}0.01 \pm 0.00$} & \textcolor{green}{$\phantom{-}0.01 \pm 0.00$} & \textcolor{green}{$\phantom{-}0.01 \pm 0.00$} & \textcolor{green}{$\phantom{-}0.01 \pm 0.00$} \\
deu-fra & $\phantom{-}0.00 \pm 0.00$ & \textcolor{red}{$-0.10 \pm 0.06$} & $\phantom{-}0.00 \pm 0.00$ & \textcolor{green}{$\phantom{-}0.01 \pm 0.00$} & \textcolor{green}{$\phantom{-}0.01 \pm 0.00$} & \textcolor{green}{$\phantom{-}0.01 \pm 0.00$} & \textcolor{green}{$\phantom{-}0.01 \pm 0.00$} \\

\hline
\addlinespace
\multicolumn{8}{c}{\textbf{Group 6: Languages of Different Scripts, Non-Aligned}} \\
\midrule
spa-rus & $\phantom{-}0.00 \pm 0.00$ & \textcolor{red}{$-0.12 \pm 0.04$} & $\phantom{-}0.00 \pm 0.00$ & $\phantom{-}0.00 \pm 0.00$ & $\phantom{-}0.00 \pm 0.00$ & $\phantom{-}0.00 \pm 0.00$ & $\phantom{-}0.00 \pm 0.00$ \\
spa-kas & $\phantom{-}0.00 \pm 0.01$ & $-0.02 \pm 0.02$ & $-0.02 \pm 0.02$ & \textcolor{green}{$\phantom{-}0.06 \pm 0.01$} & \textcolor{green}{$\phantom{-}0.03 \pm 0.01$} & \textcolor{green}{$\phantom{-}0.06 \pm 0.03$} & \textcolor{green}{$\phantom{-}0.04 \pm 0.01$} \\
heb-fra & \textcolor{green}{$\phantom{-}0.03 \pm 0.00$} & \textcolor{red}{$-0.29 \pm 0.10$} & $\phantom{-}0.01 \pm 0.02$ & \textcolor{green}{$\phantom{-}0.03 \pm 0.01$} & \textcolor{green}{$\phantom{-}0.03 \pm 0.00$} & \textcolor{green}{$\phantom{-}0.03 \pm 0.01$} & \textcolor{green}{$\phantom{-}0.03 \pm 0.01$} \\
hin-zho & \textcolor{green}{$\phantom{-}0.03 \pm 0.01$} & \textcolor{red}{$-0.37 \pm 0.06$} & $-0.01 \pm 0.04$ & \textcolor{green}{$\phantom{-}0.03 \pm 0.02$} & \textcolor{green}{$\phantom{-}0.03 \pm 0.02$} & $\phantom{-}0.02 \pm 0.03$ & $\phantom{-}0.02 \pm 0.03$ \\

\hline
\addlinespace
\multicolumn{8}{c}{\textbf{Group 7: Languages of Different Scripts, Aligned}} \\
\midrule
deu-heb & $\phantom{-}0.00 \pm 0.00$ & \textcolor{red}{$-0.21 \pm 0.07$} & $\phantom{-}0.00 \pm 0.00$ & $\phantom{-}0.00 \pm 0.00$ & $\phantom{-}0.00 \pm 0.00$ & $\phantom{-}0.00 \pm 0.00$ & $\phantom{-}0.00 \pm 0.00$ \\
kaz-hin & \textcolor{green}{$\phantom{-}0.04 \pm 0.01$} & \textcolor{red}{$-0.39 \pm 0.16$} & $\phantom{-}0.01 \pm 0.04$ & \textcolor{green}{$\phantom{-}0.07 \pm 0.01$} & \textcolor{green}{$\phantom{-}0.07 \pm 0.01$} & \textcolor{green}{$\phantom{-}0.07 \pm 0.01$} & \textcolor{green}{$\phantom{-}0.06 \pm 0.01$} \\
jpn-rus & $\phantom{-}0.01 \pm 0.01$ & \textcolor{red}{$-0.26 \pm 0.04$} & $\phantom{-}0.00 \pm 0.02$ & \textcolor{green}{$\phantom{-}0.03 \pm 0.01$} & \textcolor{green}{$\phantom{-}0.03 \pm 0.01$} & \textcolor{green}{$\phantom{-}0.02 \pm 0.01$} & \textcolor{green}{$\phantom{-}0.03 \pm 0.00$} \\
kor-fra & \textcolor{green}{$\phantom{-}0.04 \pm 0.00$} & \textcolor{red}{$-0.50 \pm 0.17$} & $\phantom{-}0.01 \pm 0.03$ & \textcolor{green}{$\phantom{-}0.05 \pm 0.01$} & \textcolor{green}{$\phantom{-}0.05 \pm 0.01$} & $\phantom{-}0.03 \pm 0.04$ & $\phantom{-}0.02 \pm 0.03$ \\
\bottomrule
\end{tabular}%
}
\caption{R@10 $\Delta$ from baseline (\xlmr), averaged over all aligned models and grouped by loss configuration. Color coding indicates the averaged performance change magnitude: \textcolor{red}{red} for degradations, \textcolor{green}{green} for gains (adjusted for standard deviation).}
\label{tab:delta-r10}
\end{table*}

\newpage
\begin{table*}[t]
  \centering
  \small
  \resizebox{\textwidth}{!}{  \begin{tabular}{lccccccc}
    \toprule
    Model & S0T0M1 & S0T1M0 & S0T1M1 & S1T0M0 & S1T0M1 & S1T1M0 & S1T1M1 \\
    \midrule
    \multicolumn{8}{c}{\textbf{POS Tagging}} \\
    \midrule
    \xlmrdeen & $ \phantom{-}.00 $ & $ \phantom{-}.00 $ & $ \textcolor{red}{-.01} $ & $ \textcolor{red}{-.10} $ & $ \textcolor{red}{-.02} $ & $ \textcolor{red}{-.78} $ & $ \textcolor{red}{-.03} $ \\
    \xlmrenkk & $ \textcolor{red}{-.02} $ & $ \textcolor{red}{-.03} $ & $ \textcolor{red}{-.02} $ & $ \textcolor{red}{-.02} $ & $ \textcolor{red}{-.06} $ & $ \textcolor{red}{-.08} $ & $ \textcolor{red}{-.10} $ \\
    \xlmrjakk & $ \phantom{-}.08 $ & $ \phantom{-}.04 $ & $ \phantom{-}.08 $ & $ \textcolor{red}{-.01} $ & $ \phantom{-}.08 $ & $ \phantom{-}.00 $ & $ \phantom{-}.10 $ \\
    \xlmrkkko & $ \textcolor{red}{-.01} $ & $ \textcolor{red}{-.02} $ & $ \textcolor{red}{-.02} $ & $ \textcolor{red}{-.01} $ & $ \textcolor{red}{-.08} $ & $ \textcolor{red}{-.01} $ & $ \textcolor{red}{-.07} $ \\
    \midrule
    \multicolumn{8}{c}{\textbf{Sentence Classification}} \\
    \midrule
    \xlmrdeen & $ \phantom{-}.03 $ & $ \textcolor{red}{-.05} $ & $ \textcolor{red}{-.03} $ & $ \phantom{-}.12 $ & $ \textcolor{red}{-.03} $ & $ \phantom{-}.12 $ & $ \textcolor{red}{-.03} $ \\
    \xlmrenkk & $ \phantom{-}.12 $ & $ \textcolor{red}{-.03} $ & $ \phantom{-}.11 $ & $ \phantom{-}.18 $ & $ \phantom{-}.03 $ & $ \phantom{-}.11 $ & $ \phantom{-}.01 $ \\
    \xlmrjakk & $ \phantom{-}.06 $ & $ \phantom{-}.01 $ & $ \textcolor{red}{-.02} $ & $ \phantom{-}.13 $ & $ \phantom{-}.01 $ & $ \phantom{-}.04 $ & $ \phantom{-}.07 $ \\
    \xlmrkkko & $ \phantom{-}.13 $ & $ \phantom{-}.02 $ & $ \phantom{-}.01 $ & $ \phantom{-}.12 $ & $ \phantom{-}.02 $ & $ \phantom{-}.05 $ & $ \phantom{-}.02 $ \\
    \bottomrule
  \end{tabular}}
  \caption{Cosine similarity between task gradients for \textsc{POS Tagging} (\textbf{top}) and \textsc{Sentence Classification} (\textbf{bottom}) and respective alignment objective by model (aligned and then fine-tuned on the task). \textcolor{red}{Red} indicates negative similarity.}
  \label{tab:align_task_grad}
\end{table*}

\newpage
\begin{table*}[ht!]
\centering
\small
\resizebox{\textwidth}{!}{
\begin{tabular}{lccccccc}
\toprule
Lang Pair & S0T0M1 & S0T1M0 & S0T1M1 & S1T0M0 & S1T0M1 & S1T1M0 & S1T1M1 \\
\midrule
\multicolumn{8}{c}{\textbf{Group 1: Aligned Languages With English}} \\
\midrule
deu-eng & 
$\phantom{-}.00 \pm .00$ & 
\textcolor{red}{$-.01 \pm .00$} & 
$\phantom{-}.00 \pm .00$ & 
\textcolor{green}{$\phantom{-}.01 \pm .00$} & 
\textcolor{green}{$\phantom{-}.01 \pm .00$} & 
\textcolor{green}{$\phantom{-}.01 \pm .00$} & 
\textcolor{green}{$\phantom{-}.01 \pm .00$} \\
kaz-eng & 
$\phantom{-}.00 \pm .00$ & 
\textcolor{red}{$-.03 \pm .01$} & 
$\phantom{-}.00 \pm .00$ & 
\textcolor{green}{$\phantom{-}.01 \pm .00$} & 
\textcolor{green}{$\phantom{-}.01 \pm .00$} & 
\textcolor{green}{$\phantom{-}.01 \pm .00$} & 
\textcolor{green}{$\phantom{-}.01 \pm .00$} \\
kor-eng & 
$\phantom{-}.00 \pm .00$ & 
\textcolor{red}{$-.02 \pm .01$} & 
$\phantom{-}.00 \pm .00$ & 
\textcolor{green}{$\phantom{-}.01 \pm .00$} & 
\textcolor{green}{$\phantom{-}.01 \pm .00$} & 
\textcolor{green}{$\phantom{-}.01 \pm .00$} & 
\textcolor{green}{$\phantom{-}.01 \pm .00$} \\
jpn-eng & 
$\phantom{-}.00 \pm .00$ & 
\textcolor{red}{$-.02 \pm .00$} & 
\textcolor{green}{$\phantom{-}.01 \pm .00$} & 
\textcolor{green}{$\phantom{-}.01 \pm .00$} & 
\textcolor{green}{$\phantom{-}.01 \pm .00$} & 
\textcolor{green}{$\phantom{-}.01 \pm .00$} & 
\textcolor{green}{$\phantom{-}.01 \pm .00$} \\
\hline
\addlinespace
\multicolumn{8}{c}{\textbf{Group 2: Aligned Languages Without English}} \\
\midrule
jpn-kaz & 
$\phantom{-}.00 \pm .00$ & 
\textcolor{red}{$-.02 \pm .01$} & 
$\phantom{-}.00 \pm .00$ & 
\textcolor{green}{$\phantom{-}.01 \pm .00$} & 
\textcolor{green}{$\phantom{-}.01 \pm .00$} & 
\textcolor{green}{$\phantom{-}.01 \pm .00$} & 
\textcolor{green}{$\phantom{-}.01 \pm .00$} \\
kor-kaz & 
$\phantom{-}.00 \pm .00$ & 
$\phantom{-}.01 \pm .01$ & 
$\phantom{-}.00 \pm .00$ &
\textcolor{green}{$\phantom{-}.01 \pm .00$} & 
\textcolor{green}{$\phantom{-}.01 \pm .00$} & 
\textcolor{green}{$\phantom{-}.01 \pm .00$} & 
\textcolor{green}{$\phantom{-}.01 \pm .00$} \\
\hline
\addlinespace
\multicolumn{8}{c}{\textbf{Group 3: Other Languages With English}} \\
\midrule
hin-eng & 
$\phantom{-}.00 \pm .00$ & 
\textcolor{red}{$-.02 \pm .01$} & 
$\phantom{-}.00 \pm .00$ & 
\textcolor{green}{$\phantom{-}.01 \pm .00$} & 
\textcolor{green}{$\phantom{-}.01 \pm .00$} & 
\textcolor{green}{$\phantom{-}.01 \pm .00$} & 
\textcolor{green}{$\phantom{-}.01 \pm .00$} \\
zho-eng &
$\phantom{-}.00 \pm .00$ & 
$\phantom{-}.01 \pm .01$ & 
$\phantom{-}.00 \pm .00$ & 
\textcolor{green}{$\phantom{-}.01 \pm .00$} & 
\textcolor{green}{$\phantom{-}.01 \pm .00$} & 
\textcolor{green}{$\phantom{-}.01 \pm .00$} & 
\textcolor{green}{$\phantom{-}.01 \pm .00$} \\
spa-eng & 
$\phantom{-}.00 \pm .01$ & 
$\phantom{-}.00 \pm .00$ & 
$\phantom{-}.00 \pm .00$ & 
\textcolor{green}{$\phantom{-}.01 \pm .00$} & 
\textcolor{green}{$\phantom{-}.01 \pm .00$} & 
\textcolor{green}{$\phantom{-}.01 \pm .00$} & 
\textcolor{green}{$\phantom{-}.01 \pm .00$} \\
rus-eng & 
$\phantom{-}.00 \pm .00$ & 
$\phantom{-}.01 \pm .01$ & 
$\phantom{-}.00 \pm .00$ & 
\textcolor{green}{$\phantom{-}.01 \pm .00$} & 
\textcolor{green}{$\phantom{-}.01 \pm .00$} & 
\textcolor{green}{$\phantom{-}.01 \pm .00$} & 
\textcolor{green}{$\phantom{-}.01 \pm .00$} \\
\hline
\addlinespace
\multicolumn{8}{c}{\textbf{Group 4: Languages of the Same Script, Non-Aligned}} \\
\midrule
lit-spa & 
$\phantom{-}.00 \pm .00$ & 
\textcolor{red}{$-.01 \pm .00$} & 
$\phantom{-}.00 \pm .00$ & 
\textcolor{green}{$\phantom{-}.01 \pm .00$} & 
\textcolor{green}{$\phantom{-}.01 \pm .00$} & 
\textcolor{green}{$\phantom{-}.01 \pm .00$} & 
\textcolor{green}{$\phantom{-}.01 \pm .00$} \\
hin-kas & 
$\phantom{-}.00 \pm .00$ & 
\textcolor{red}{$-.08 \pm .05$} & 
$\phantom{-}.00 \pm .00$ & 
\textcolor{green}{$\phantom{-}.02 \pm .01$} & 
\textcolor{green}{$\phantom{-}.01 \pm .00$} & 
$\phantom{-}.01 \pm .01$ & 
$\phantom{-}.01 \pm .01$ \\
spa-fra & 
$\phantom{-}.00 \pm .00$ & 
$\phantom{-}.00 \pm .00$ & 
$\phantom{-}.00 \pm .00$ & 
\textcolor{green}{$\phantom{-}.01 \pm .00$} & 
$\phantom{-}.00 \pm .00$ & 
$\phantom{-}.00 \pm .00$ & 
\textcolor{green}{$\phantom{-}.01 \pm .00$} \\
rus-bul & 
$\phantom{-}.00 \pm .00$ & 
$\phantom{-}.00 \pm .00$ & 
$\phantom{-}.00 \pm .00$ & 
\textcolor{green}{$\phantom{-}.01 \pm .00$} & 
\textcolor{green}{$\phantom{-}.01 \pm .00$} & 
\textcolor{green}{$\phantom{-}.01 \pm .00$} & 
\textcolor{green}{$\phantom{-}.01 \pm .00$} \\
\hline
\addlinespace
\multicolumn{8}{c}{\textbf{Group 5: Languages of the Same Script, Aligned}} \\
\midrule
kaz-bul & 
$\phantom{-}.00 \pm .00$ & 
\textcolor{red}{$-.02 \pm .01$} & 
\textcolor{green}{$\phantom{-}.01 \pm .00$} & 
\textcolor{green}{$\phantom{-}.01 \pm .00$} & 
\textcolor{green}{$\phantom{-}.01 \pm .00$} & 
\textcolor{green}{$\phantom{-}.01 \pm .00$} & 
\textcolor{green}{$\phantom{-}.01 \pm .00$} \\
kaz-rus & 
$\phantom{-}.00 \pm .00$ & 
\textcolor{red}{$-.02 \pm .01$} & 
\textcolor{green}{$\phantom{-}.01 \pm .00$} & 
\textcolor{green}{$\phantom{-}.01 \pm .00$} & 
\textcolor{green}{$\phantom{-}.01 \pm .00$} & 
\textcolor{green}{$\phantom{-}.01 \pm .00$} & 
\textcolor{green}{$\phantom{-}.01 \pm .00$} \\
deu-spa & 
$\phantom{-}.00 \pm .00$ & 
$\phantom{-}.00 \pm .00$ & 
$\phantom{-}.00 \pm .00$ & 
\textcolor{green}{$\phantom{-}.01 \pm .00$} & 
\textcolor{green}{$\phantom{-}.01 \pm .00$} & 
\textcolor{green}{$\phantom{-}.01 \pm .00$} & 
\textcolor{green}{$\phantom{-}.01 \pm .00$} \\
deu-fra & 
$\phantom{-}.00 \pm .00$ & 
$\phantom{-}.00 \pm .00$ & 
$\phantom{-}.00 \pm .00$ & 
\textcolor{green}{$\phantom{-}.01 \pm .00$} & 
\textcolor{green}{$\phantom{-}.01 \pm .00$} & 
\textcolor{green}{$\phantom{-}.01 \pm .00$} & 
\textcolor{green}{$\phantom{-}.01 \pm .00$} \\
\hline
\addlinespace
\multicolumn{8}{c}{\textbf{Group 6: Languages of Different Scripts, Non-Aligned}} \\
\midrule
spa-rus & 
$\phantom{-}.00 \pm .00$ & 
\textcolor{red}{$-.01 \pm .00$} & 
$\phantom{-}.00 \pm .00$ & 
\textcolor{green}{$\phantom{-}.01 \pm .00$} &  
\textcolor{green}{$\phantom{-}.01 \pm .00$} &  
\textcolor{green}{$\phantom{-}.01 \pm .00$} & 
\textcolor{green}{$\phantom{-}.01 \pm .00$} \\
spa-kas & 
\textcolor{green}{$\phantom{-}.01 \pm .00$} & 
\textcolor{red}{$-.10 \pm .06$} & 
$\phantom{-}.00 \pm .01$ & 
\textcolor{green}{$\phantom{-}.03 \pm .01$} & 
\textcolor{green}{$\phantom{-}.02 \pm .00$} & 
\textcolor{green}{$\phantom{-}.02 \pm .00$} & 
\textcolor{green}{$\phantom{-}.02 \pm .01$} \\
heb-fra & 
\textcolor{green}{$\phantom{-}.01 \pm .00$} & 
$\phantom{-}.01 \pm .01$ & 
\textcolor{green}{$\phantom{-}.01 \pm .00$} & 
\textcolor{green}{$\phantom{-}.01 \pm .00$} & 
\textcolor{green}{$\phantom{-}.01 \pm .00$} &  
\textcolor{green}{$\phantom{-}.01 \pm .00$} & 
\textcolor{green}{$\phantom{-}.01 \pm .00$} \\
hin-zho & 
$\phantom{-}.00 \pm .00$ & 
$\phantom{-}.01 \pm .01$ & 
\textcolor{green}{$\phantom{-}.01 \pm .00$} & 
\textcolor{green}{$\phantom{-}.01 \pm .00$} & 
\textcolor{green}{$\phantom{-}.01 \pm .00$} & 
\textcolor{green}{$\phantom{-}.01 \pm .00$} & 
\textcolor{green}{$\phantom{-}.01 \pm .00$} \\
\hline
\addlinespace
\multicolumn{8}{c}{\textbf{Group 7: Languages of Different Scripts, Aligned}} \\
\midrule
deu-heb & 
\textcolor{green}{$\phantom{-}.01 \pm .00$} & 
$\phantom{-}.01 \pm .01$ & 
\textcolor{green}{$\phantom{-}.01 \pm .00$} &  
\textcolor{green}{$\phantom{-}.01 \pm .00$} &  
\textcolor{green}{$\phantom{-}.01 \pm .00$} & 
\textcolor{green}{$\phantom{-}.01 \pm .00$} &  
\textcolor{green}{$\phantom{-}.01 \pm .00$} \\
kaz-hin & 
$\phantom{-}.00 \pm .00$ &  
$\phantom{-}.01 \pm .01$ &     
$\phantom{-}.00 \pm .00$ &    
\textcolor{green}{$\phantom{-}.01 \pm .00$} &    
$\phantom{-}.00 \pm .00$ &   
$\phantom{-}.00 \pm .00$ &   
\textcolor{green}{$\phantom{-}.01 \pm .00$} \\  
jpn-rus &
$\phantom{-}.00 \pm .00$ &  
$\phantom{-}.01 \pm .01$ &   
\textcolor{green}{$\phantom{-}.01 \pm .00$} &  
\textcolor{green}{$\phantom{-}.01 \pm .00$} & 
\textcolor{green}{$\phantom{-}.01 \pm .00$} & 
\textcolor{green}{$\phantom{-}.01 \pm .00$} &   
\textcolor{green}{$\phantom{-}.01 \pm .00$} \\  
kor-fra &
\textcolor{green}{$\phantom{-}.01 \pm .00$} &     
$\phantom{-}.01 \pm .01$ &          
\textcolor{green}{$\phantom{-}.01 \pm .00$} &         
\textcolor{green}{$\phantom{-}.01 \pm .00$} &     
\textcolor{green}{$\phantom{-}.01 \pm .00$} &     
\textcolor{green}{$\phantom{-}.01 \pm .00$} &     
\textcolor{green}{$\phantom{-}.01 \pm .00$} \\  
\bottomrule
\end{tabular}}
\caption{Cosine similarity $\Delta$ from \xlmr, averaged over all aligned models and grouped by loss configuration. 
Color coding indicates the averaged similarity change magnitude: 
\textcolor{red}{red} for decreased similarity (accounting for standard deviation), \textcolor{green}{green} for gains from $+.01$ and above, black for negligible improvements or inconclusive measurements.}

\label{tab:st_similarity_differences}
\end{table*}

\end{document}